\documentclass[sigconf]{acmart}
\usepackage{comment}

\usepackage{algorithm2e}
\usepackage{subcaption}
\usepackage{listings}
\RestyleAlgo{ruled}
\SetKwComment{Comment}{/* }{ */}

\AtBeginDocument{%
  }

\setcopyright{cc}
\acmDOI{}
\acmConference{LIACS}{2025}{Leiden}

\acmISBN{}

\begin{document}


%\title{Mirror Mode: Agents Imitating Player's Strategy}

\title[Mirror Mode in Fire Emblem]{Mirror Mode in Fire Emblem:\\ Beating Players at their own Game\\ with Imitation and Reinforcement Learning}

% %\title{Mirror Mode:\\ Applying Reinforcement Learning and Imitation Learning\\ to Use a Player’s Strategy Against Them\\ in Turn-Based Strategy Games}


%\begin{comment}

\author{Yanna Elizabeth Smid}
\email{yanna.e.smid@gmail.com} 
\orcid{0009-0004-4095-5749}
\affiliation{%
  \institution{LIACS, Leiden University}
  \city{Leiden}
  \country{The Netherlands}
}

\author{Peter van der Putten}
\email{p.w.h.van.der.putten@liacs.leidenuniv.nl}
\orcid{0000-0002-6507-6896}
\affiliation{%
  \institution{LIACS, Leiden University}
  \city{Leiden}
  \country{The Netherlands}
}

\author{Aske Plaat}
\email{a.plaat@liacs.leidenuniv.nl}
\orcid{0000-0001-7202-3322}
\affiliation{%
  \institution{LIACS, Leiden University}
  \city{Leiden}
  \country{The Netherlands}
}

%\end{comment}

%\renewcommand{\shortauthors}{Y. E. Smid et al.}

\begin{abstract}
Enemy strategies in turn-based games should be surprising and unpredictable. This study introduces Mirror Mode, a new game mode where the enemy AI mimics the personal strategy of a player to challenge them to keep changing their gameplay. A simplified version of the Nintendo strategy video game Fire Emblem Heroes has been built in Unity, with a Standard Mode and a Mirror Mode. Our first set of experiments find a suitable model for the task to imitate player demonstrations, using  Reinforcement Learning and Imitation Learning: combining Generative Adversarial Imitation Learning, Behavioral Cloning, and Proximal Policy Optimization. The second set of experiments evaluates the constructed model with player tests, where models are trained on  demonstrations provided by participants. The gameplay of the participants indicates  good imitation in defensive behavior, but not in offensive strategies. Participant's surveys indicated that they recognized their own retreating tactics, and resulted in an overall higher player-satisfaction for Mirror Mode. 
Refining the model further may improve imitation quality and increase player's satisfaction, especially when players face their own strategies.
The full code and survey results are stored at: \url{https://github.com/YannaSmid/MirrorMode}.
\end{abstract}

\begin{comment}
\begin{teaserfigure}
  \includegraphics[width=\textwidth]{sampleteaser}
  \caption{Seattle Mariners at Spring Training, 2010.}
  \Description{Enjoying the baseball game from the third-base
  seats. Ichiro Suzuki preparing to bat.}
  \label{fig:teaser}
\end{teaserfigure}
\end{comment}


\maketitle

\section{Introduction}
In video games, non-playable character (NPC) behavior has relied on artificial intelligence (AI) algorithms for decades \cite{tian2024}. Now, with the quick advancements made in AI, new possibilities are found to enhance the behavior of NPCs, to increase the quality of a video game. 
%Artificial Intelligence (AI) in Game development is a widely studied topic, providing new opportunities for a diverse range of aspects such as improving the game quality, enhancing a player's game experience, or faster level generation. In video games the behavior of non-playable characters (NPCs) are carried by AI algorithms.
NPC behavior refers to how characters in games should act and react to certain events in the game environment. Realistic NPC behavior contributes significantly to the player immersion and satisfaction of the game \cite{lee2012}. 
%Ideally, each NPC should exhibit behavior that aligns with its role in the game. %For instance, a knight character might be programmed to act more aggressively in combat, while a cleric would prioritize healing or supporting allies.
Traditionally, these behavior types are handled by Finite State Machines (FSM) or Behavior Trees (BT), where each character follows a set of predefined heuristics and transitions between states based on game events \cite{Fan2023}. Nevertheless, this method of programmable behavior can result in repetitive behavior that makes NPCs predictable in their actions \cite{akram2024,lemaitre2015}. 

In strategy games, predictability in enemy tactics can have a major influence on player experience. Strategy games require tactical thinking to defeat a team of opponents, while keeping your own team alive. Statistics in 2024 have shown that the popularity of strategy games has drastically decreased in the past 9 years \cite{quantic}. This may be linked to the predictability of the enemy's action as it makes games easier to play, possibly reducing the engagement of more experienced players \cite{akram2024}. In addition to this, it is found that playing several repetitive games can cause boredom \cite{chanel2008}.  

Therefore, this study aims to address the risk of boredom in strategy video games by introducing \textit{Mirror Mode}, a new game mode where the enemy NPCs learn a strategy based on the player's strategy through Imitation Learning (IL) \cite{ross2010}. A simplified version of the mobile strategy game \textit{Fire Emblem Heroes} \cite{NintendoFEH} was developed, to apply combinations of Generative Adversarial Imitation Learning (GAIL), Behavioral Cloning (BC), and Proximal Policy Optimization (PPO), in order to train agents.
%in the Fire Emblem environment and compared to each other, including Generative Adversarial Imitation Learning (GAIL), Behavioral Cloning (BC), and Proximal Policy Optimization (PPO). 
%The algorithms are tested among different parameters to find the optimal model that is used for the player tests. Ultimately, the optimal hyperparameters are used for the model that are trained on real player's data. Tests are conducted where participants need to play the created game for five rounds against the standard enemy AI. A second test is taken where each player is up against an enemy AI trained by either their demonstrated data or that from another participant. Their game experience and satisfaction is measured for both the Standard Mode as well as the Mirror Mode via questionnaires and gameplay metrics. 
The performance of the algorithms are assessed through ablation and optimization experiments, where the best configuration of algorithms and their hyperparameters were used for further evaluations through user studies. The conducted user studies evaluated the performance of the trained agents and their imitation quality.
The research aims to answer the central research question: \textit{How will a player's game experience be influenced when NPCs imitate their strategy in a turn-based strategy game?} 

With the results of the finetuning experiments the study aims to further explore the sub-question: ``\textit{To what extent can reinforcement learning (RL) and imitation learning (IL) be applied to teach NPCs the strategy of a player in a turn-based strategy game?}''

Altogether, the study provides an innovative method of playing strategy games, and offers insights into the effectiveness of IL and RL for large discrete spaces, to imitate individual player strategies. The key contributions of this study are as follows.
\begin{itemize}
    \item We implement a new strategy game   where agents are trained using Behavioral Cloning, Generative Adversarial Imitation Learning, and Reinforcement Learning.
    \item We report the first  evidence of behavior cloning in a player's strategy in a turn-based game.
    \item Users report higher satisfaction with the game through the increased interaction. 
\end{itemize}
% Accordingly, the following hypothesis will be tested: \textit{The game experience of a player will be positively influenced after the enemy AI has imitated their strategy, as the player will be more engaged and more satisfied with the game.}

% The remainder of this thesis will discuss the approach and development of addressing these setup questions and hypothesis. First, Section \ref{sec:relatedwork} reviews prior research relevant to this study. This is followed by Section \ref{sec:background}, which briefly explains the key concepts needed for this research. Before going deeper into the aspects of the study, the setup for the game and virtual environment is given in Section \ref{sec:setup}. After this, Section \ref{sec:FEunity} describes the implementation of the Fire Emblem strategy game variant used as a test environment. Subsequently, the experiments including the corresponding results will be discussed into two sections. Section \ref{sec:modeltesting} presents the training and configuration of the imitation models, followed by Section \ref{sec:playertests} which explains the evaluation of the resulted imitation model through player tests. The results from both experiments will be discussed in Section \ref{sec:discussion}, as well as the limitations of this study and possible future work. Finally, Section \ref{sec:conclusion} concludes the thesis by summarizing the findings and answering the aforementioned questions and hypothesis.

\section{Related Work} \label{sec:relatedwork}
Maintaining a balance between the skills of a player and the difficulty of the game is a key principle in game design, to influence the player's engagement and satisfaction.
According to Csíkszentmihályi's flow theory, an individual is most engaged when the challenge of an activity is well aligned to their skill level, creating an optimal state of focus and satisfaction \cite{csíkszentmihályi, huang2023}. 

However, game experience is also shaped by the motivation to play a game, which differs for each player. Players who focus primarily on achieving in-game objectives tend to be more satisfied with smooth and accessible gameplay, while those who play for enjoyment are more engaged by challenging and exciting gameplay \cite{chang2020}. This highlights the difficulty of creating a single game experience that satisfies all types of players.
Many game developers provide adjustable difficulty settings, allowing players to match the challenge to their abilities as they play. However, frequently adjusting the difficulty can break immersion and disrupt the player's sense of flow, making it harder for them to stay engaged over time \cite{huang2023}. Researchers are therefore finding a more suitable method for adapting difficulty in video games, with smarter AI algorithms.

\subsection{Adaptive Gameplay with AI}

Building onto the difficulty settings of video games, AI offers new methods in order to maintain a player's satisfaction.

Early work by Sanchez-Ruiz et al. explored NPC adaptation in the turn-based strategy game Call to Power2, using an ontological approach \cite{sanchezruiz2008}. Agents retrieve actions by looking up similar states in a library, with stored previously used tactics. 
While it proved effective in improving decision-making speed and accuracy, the approach was limited in adapting to entirely new conditions and involved computational complexity due to its ontology-based reasoning. 

Research by Akram et al. has shown that the player satisfaction has improved after the implementation of AI driven mechanics for the animation of NPCs, to improve the realism in their behavior. They suggest that further improvement on the game satisfaction can include using AI for the adaptability of NPCs on player's gaming behavior \cite{akram2024}. 

%In addition to this, Modern Game AI has already proven its success in record breaking games such as 2020s game of the year, The Last of Us 2, where enemies are more aware of their environment and stay alerted when they found an ally dead \cite{GOTY, interviewlastofus2}. Therefore, implementing new AI methods to improve the adaptability of NPCs and reduce the predictability of their actions, can potentially help maintaining the satisfaction and flow of a game.

%However, for a more advanced enemy behavior, different approaches need to be found to create agents that are adaptable to the environment. RL in the context of video games still has its limitations, due to the heavy computational demands, arbitrary reward systems, and slow learning curves. This often makes RL a less desirable method for complex game environments. 
%However, video games are considerably a complex environment, RPG strategy games are considerably a complex environment, making it challenging to specify a suitable game state for the RL algorithms, and require a high degree of adaptability in the game \cite{zare2023}. 

%RL and Video Games share a mutual interest, where RL offers new possibilities for creating intelligent game agents, while video games provide a practical environment for testing and developing RL techniques \cite{Fan2023}. Over the years, RL has been applied to NPC behavior in games to make them more complex and adaptable. 

Whilst several studies have explored AI's potentials for NPC behavior adaptation, little research has built on the idea to train agents based on real-time player strategies to improve the tactics adaptation in strategy video games. Strategy video game environments are considerably complex, making it challenging to train computationally demanding AI algorithms such as RL \cite{zare2023}.

\subsection{RL and IL for Adaptive Gameplay}

% goes back decadeds
Researchers explored a wider range of possibilites of RL for adaptive behavior. OpenAI Five was the first AI system that was able to defeat professional players in the multiplayer real-time strategy game Dota 2 \cite{openai2019dota}. Through deep RL and self-play, the algorithm succeeded in defeating the world champions in the game, marking it a great success for the use of RL in real-time strategy games.
Continuing on this work, OpenAI demonstrated the adaptability of agents while letting them play a game of hide-and-seek, through self-play \cite{baker2020}. The research applied PPO and Generalized Advantage Estimation to optimize their policy. It was found that through self-play, agents learned to develop counter-strategies to earlier discovered in-game strategies. 

%Self-play enables agents to continuously adapt their policy by playing against agents using the same policy that gets updated iteratively. It was found that through self-play, agents learned to develop counter-strategies to earlier discovered in-game strategies. Although the hide-and-seek environment differs from turn-based strategy games, this research shows how RL agents can evolve adaptive behaviors based on in-game interactions and strongly motivates the use of RL and self-play for adaptable behavior.

Amato and Shani further investigate adaptive strategy behavior for NPCs, in the turn-based strategy game Civilization \cite{amatoandshani}. They applied  Q-learning and Dyna-Q for switching between strategies given the current game situation. Their results show high potential for the use of RL to teach agents specific strategies, and the authors encourage investigating the use of more advanced RL approaches in the future. 

According to research by Zare et al., IL offers a more advanced RL approach in complex learning environments, by incorporating human demonstrations to teach agents the targeted behavior \cite{zare2023}. 

Ho and Ermon introduced Generative Adversarial Inverse Learning (GAIL), an adversarial approach that enables policy training from expert demonstrations without explicit reward systems \cite{HOandERMONGAIL}. Subsequently, Gharbi and Fennan compared GAIL with alternative IL and RL algorithms including PPO and BC, finding it highly effective at replicating complex player strategies \cite{gharbi2024}. They further suggest that combining GAIL with model-free RL methods could yield adaptive game AI capable of both imitation and responsiveness, a claim that this study is going to investigate further by incorporating GAIL with PPO.

\subsection{Research Gap}

While prior work has not examined RL and IL techniques for imitating player behavior in turn-based strategy games specifically, the aforementioned existing research highlights the potential of integrating AI in video games to produce a more adaptive and engaging game experience. The studies motivate examining the effects of PPO, BC, and GAIL combined in strategy video games. Accordingly, this study further explores the application of IL in strategic game environments to find the opportunities of creating more adaptable agents that copy strategies.

\section{Game Environment} \label{sec:FEunity}
This study developed a game environment to investigate the impact of an enemy AI that imitates player behavior. The technical setup as well as the implementation and design of the game environment is explained in this section, details on the agent AI can be found in the next section.

\subsection{Fire Emblem Heroes} \label{subsec:feheroes}

\begin{figure}[t]
    \centering

    \begin{subfigure}[b]{0.23\linewidth}
        \centering
        \includegraphics[width=\linewidth]{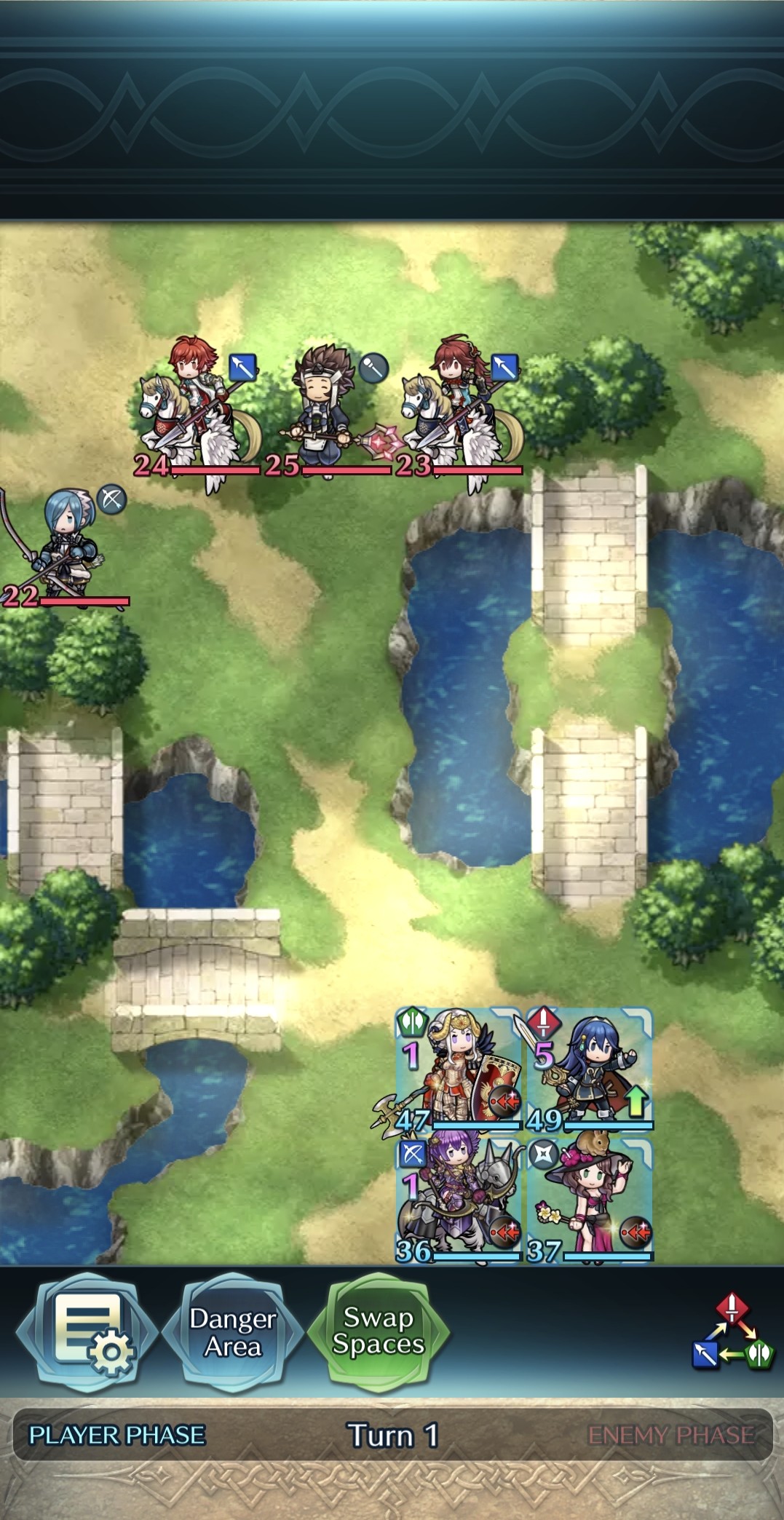}
        \caption{Start.}
        \label{fig:fehstart}
    \end{subfigure}\hfill
    \begin{subfigure}[b]{0.23\linewidth}
        \centering
        \includegraphics[width=\linewidth]{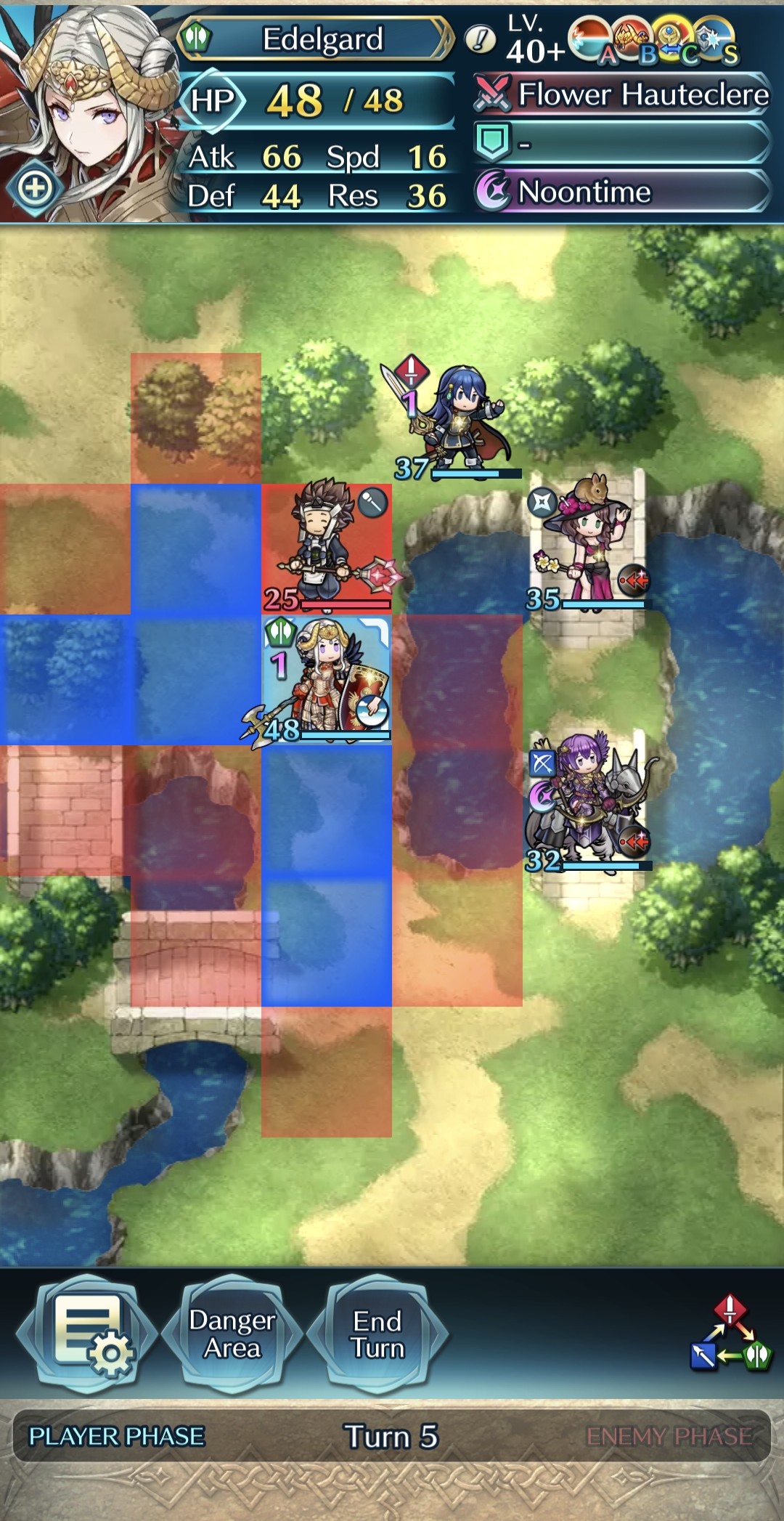}
        \caption{Ranges.}
        \label{fig:fehrange}
    \end{subfigure}\hfill
    \begin{subfigure}[b]{0.23\linewidth}
        \centering
        \includegraphics[width=\linewidth]{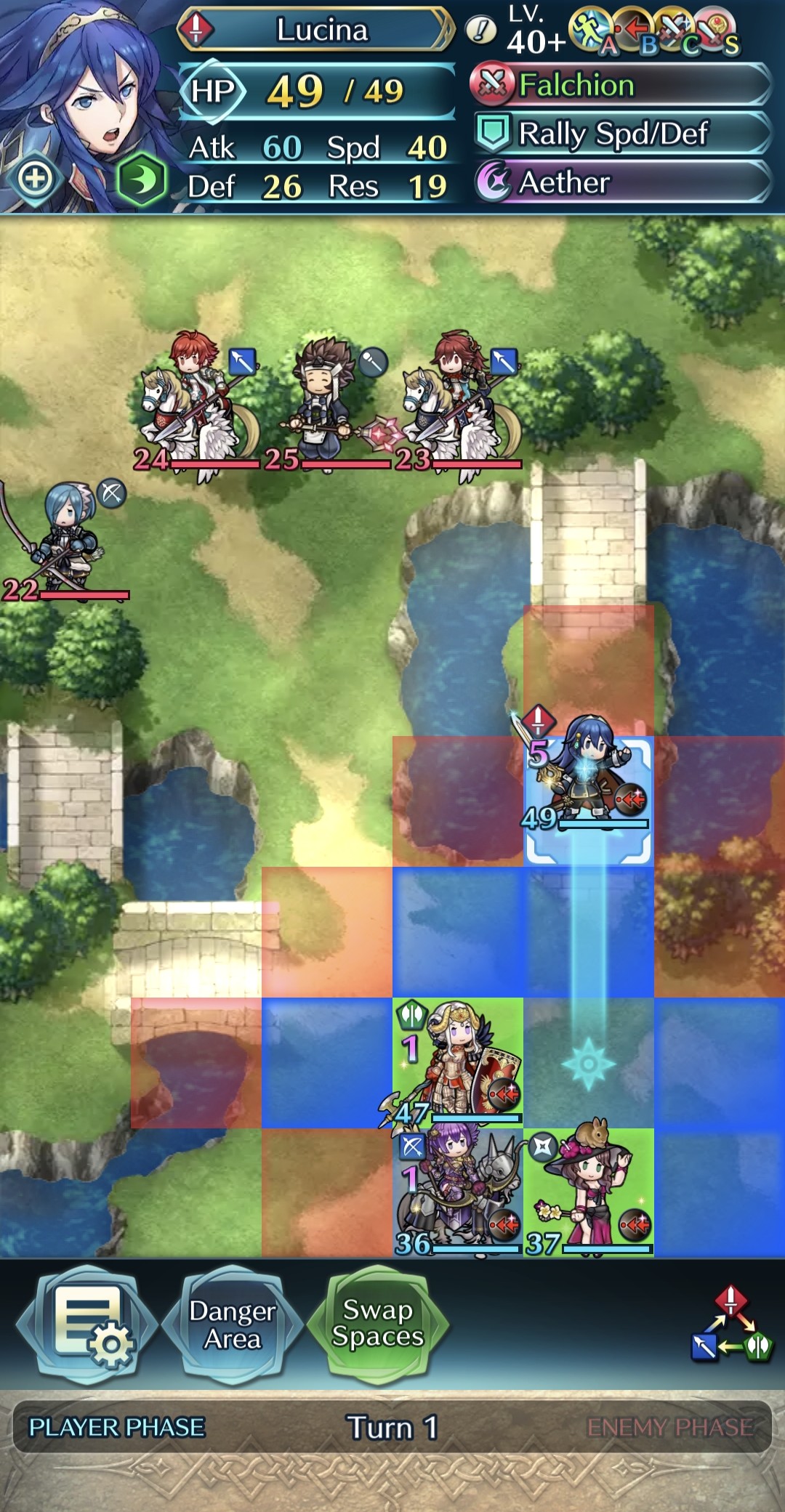}
        \caption{Move.}
        \label{fig:fehmove}
    \end{subfigure}\hfill
    \begin{subfigure}[b]{0.23\linewidth}
        \centering
        \includegraphics[width=\linewidth]{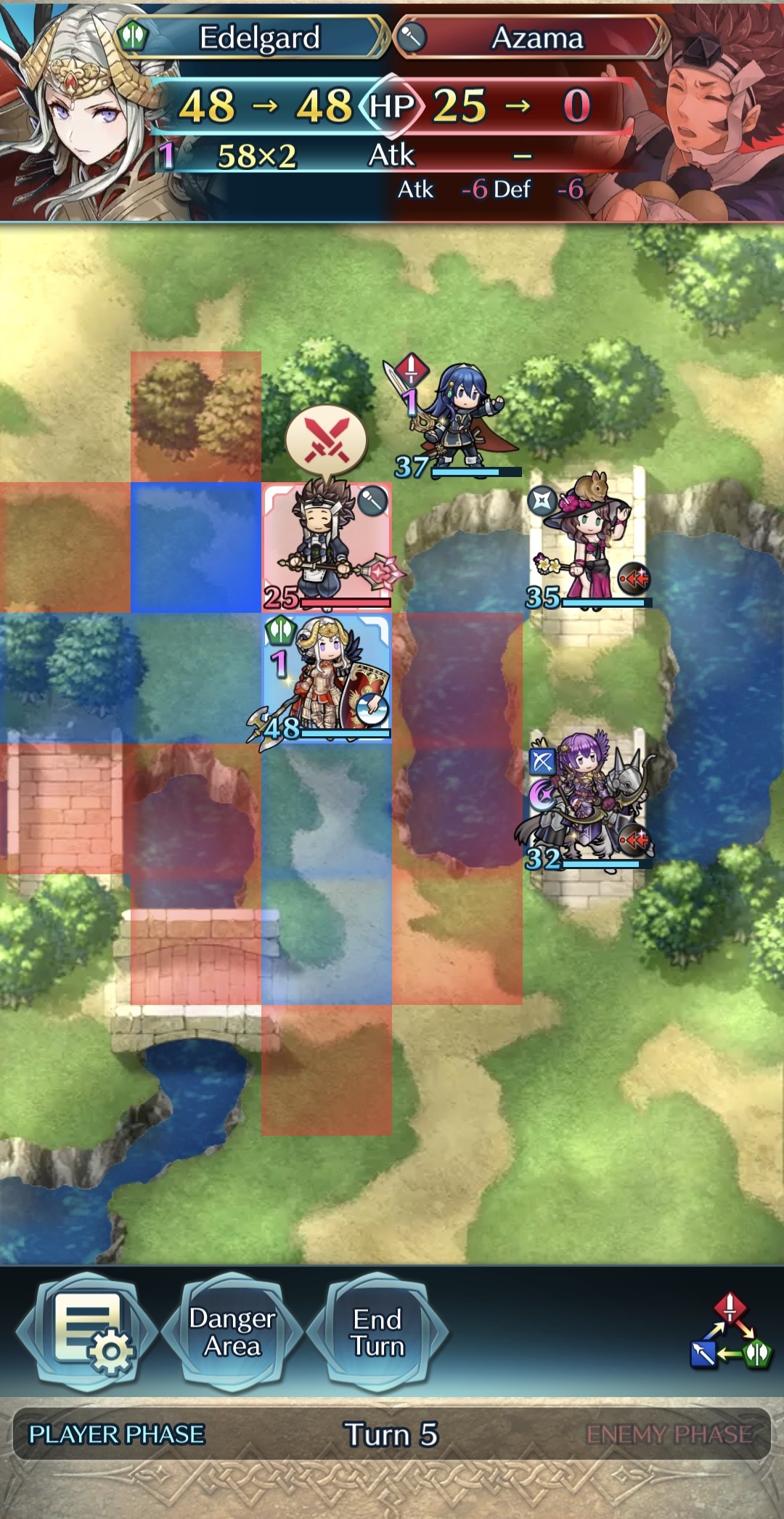}
        \caption{Combat.}
        \label{fig:fehattack}
    \end{subfigure}

    \caption{Different interface UIs from Fire Emblem Heroes by Nintendo and Intelligent Systems \cite{NintendoFEH}.}
    \label{fig:feh_figs}
\end{figure}

To study the potential of RL and IL in strategy video games, Fire Emblem Heroes is used as a layout of the typical tactical rules and designs in a regular strategy video game setting. Fire Emblem is a turn-based strategy role-playing game developed by Intelligent Systems and published by Nintendo. This research uses a simplified version of the 2017 mobile game adaption, Fire Emblem Heroes \cite{NintendoFEH}, maintaining the complex tactical thinking in a compact $6\times8$ grid-based map, such as the maps shown in Figure \ref{fig:feh_figs}.  

The game revolves around an alternating player and enemy phase. Starting with the player phase, all units are positioned on one tile on the map. Characteristics corresponding to the player's team can be recognized by  its blue color, while the enemy team is always represented by the red color. An example of the start interface can be seen in Figure \ref{fig:fehstart}.

In each phase, all surviving units in a team have one turn. One unit can be selected at a time to perform an action in their turn, which can be to move, wait, or attack. Attacks are only possible if a foe is within the attack range of a unit, and a target may counterattack if their range matches the attacker's range. Figure \ref{fig:fehrange} shows how the game presents tiles that are within attack or movement range. Tiles that are in attack range are highlighted in red, and tiles that are within movement range in blue. A unit can move to any tile highlighted in blue, as given in Figure \ref{fig:fehmove}. If an enemy stands on a tile within attack range, the tile is highlighted in brighter red to indicate that the enemy can be selected to attack. When a target is selected, the combat information appears in the top of the screen, shown in Figure \ref{fig:fehattack}.

The game ends when all units on one side are defeated. Each team contains four units, and the game starts at the player phase. 

Each unit belongs to one of four types: infantry, cavalry, flying, and heavy armor units. The unit type determines their terrain accessibility, step size, and attack and defense power named as stats. 
Units also carry one of the five weapon types: sword, lance, axe, bow, or magic. Weapons determine attack range and can apply advantages and effectiveness. Melee weapons (sword, axe, lance) have a range of one tile, while bow and magic can attack from two tiles away. 

\begin{figure}[t]
    \centering
    \includegraphics[width=0.3\linewidth]{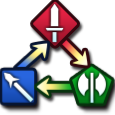}
    \caption{The weapon triangle in Fire Emblem, showing advantages/disadvantages of each melee weapon \cite{FEUIsprites}.}
    \label{fig:weapontriangle}
\end{figure}

The strategic thinking arises from the interaction of unit types, their weapons, and stats. The melee weapons are part of a weapon triangle, presented in Figure \ref{fig:weapontriangle}, similar to the "rock, paper, scissors" principle. Sword has an advantage against axe, axe against lance, and lance against sword. Boosting damage by 1.2x, or 0.8 the other way around.

Bows are highly effective against flying units, dealing $1.5\times$ extra damage. Heavy armor units have a high attack and defense, but a low resistance, making them only vulnerable to magic. Magic units have a high resistance, and are valuable to defeat the heavy armor units, but they are vulnerable to melee attacks.

Lastly, the five core stats are important to keep in mind. HP determines the hit points the unit can take. Attack calculates the damage output by the unit. Defense is subtracted from the damage of an attacker carrying a melee weapon, whereas the resistance is subtracted from the damage dealt by a magic user. The speed enables a follow-up attack when the difference in speed between attacker and target differs at least 5 points.

Originally, the game includes more types than the aforementioned unit and weapon types. In addition, the strategies are widely influenced by special abilities, assists, and skills. To maintain simplicity, this implementation solely focuses on the interaction between the four unit types, and five weapon types. 

\subsection{Experimental Setup}\label{sec:setup}
A 2D game environment is created in Unity Engine. For the integration of ML tools from Python PyTorch library with Unity, the ML-Agent toolkit has been developed by Juliani et al., including the RL and IL algorithms programmed in Python \cite{juliani2020}. The installation steps and setup instructions are followed according to ML-Agent version release 22 \cite{ML-Agent-Installation}. 

A constructed virtual environment within the same directory of the game environment allows interaction between the game environment and the ML tools in python.

The final game environment setup is posted on our \href{https://github.com/AnonymousResearcher22/MirrorModeResearch}{GitHub page}~\cite{myproject}. More detailed installation instructions and requirements are also mentioned.

\subsection{Fire Emblem Unity Scene}\label{subsec:feunityscene}
The 2D grid-based map existing of \textit{6x8} tiles, is recreated in a Unity 2D environment. The map is designed with complete symmetry across both axes. Each team starts on its own side of the map, ensuring that all tile types and movement distances are balanced and mirrored between the player and enemy units.

\begin{figure}[t]
    \centering
    \begin{subfigure}[b]{0.48\textwidth}
        \centering
        \includegraphics[width=\textwidth]{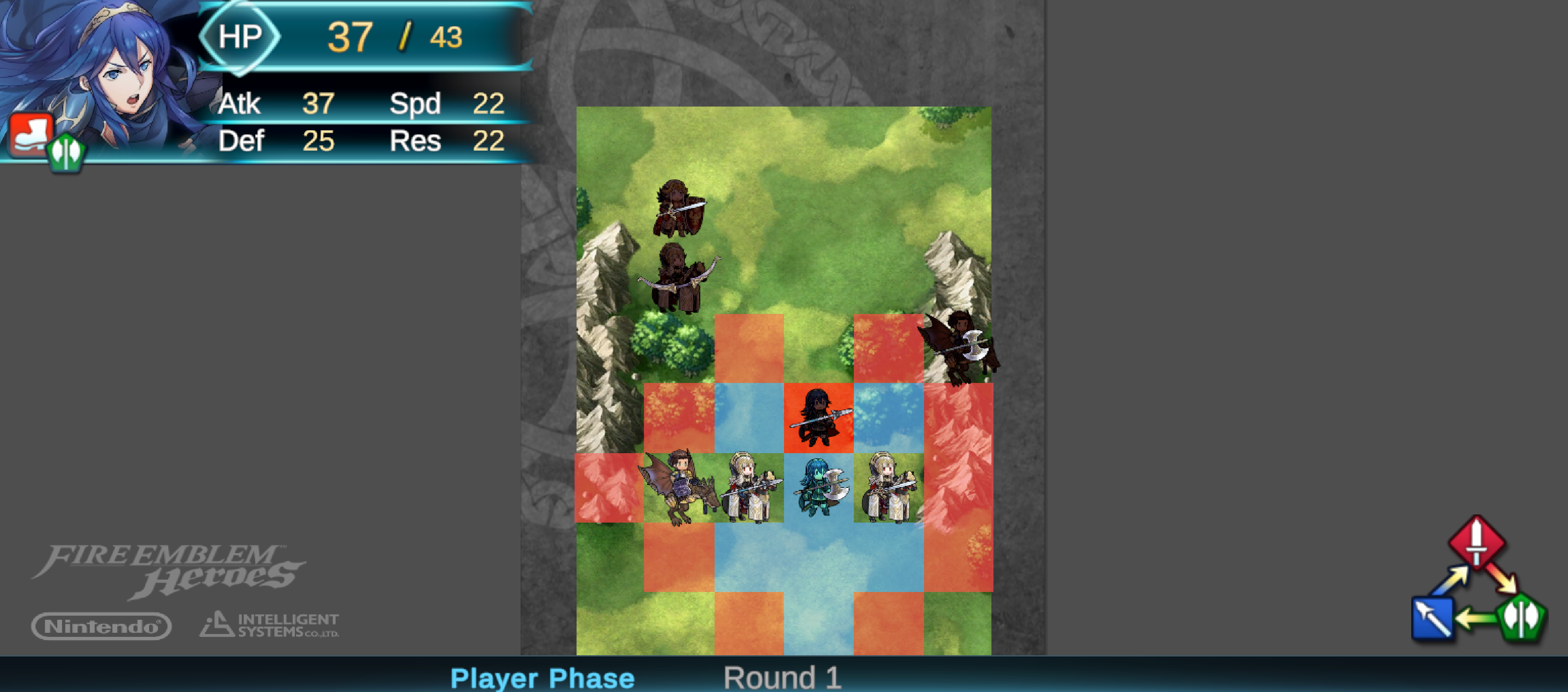}
        \caption{Ranges.}
        \label{fig:myfeh_selecting}
    \end{subfigure}
    \hfill
    \begin{subfigure}[b]{0.48\textwidth}
        \centering
        \includegraphics[width=\textwidth]{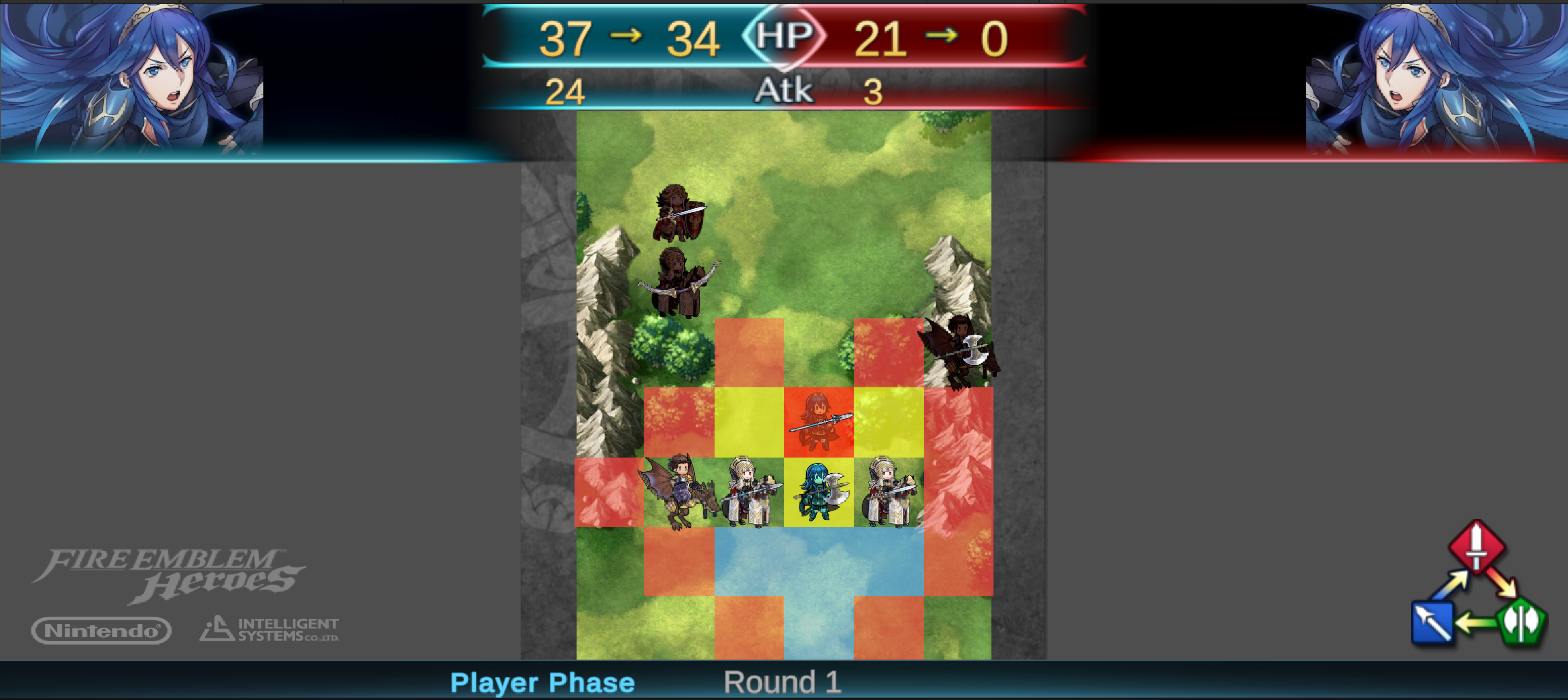}
        \caption{Combat.}
        \label{fig:myfeh_combat}
    \end{subfigure}

    \caption{End result of the implemented version of Fire Emblem Heroes for this study, with sprites from the original game, gained via Fire Emblem Heroes Wiki Fandom Community \cite{FEUIsprites, FEherolist}.}
    \label{fig:myfeh_endresult}
\end{figure}

Units are defined by a combination of movement type, weapon type, and combat stats. These variables are set public and can be assigned in the Unity Inspector tab, in the Information component of the unit. The enemy types are completely randomized each game round. For both teams, the combat stats are randomized. 
Standard Mode and Mirror Mode are each created in a separate scene. In both scenes, all visual sprites and icons are collected from publicly available Fire Emblem Fandom community resources in the game assets page \cite{FEUIsprites}, and the unit sprites from the character misc information from the heroes list \cite{FEherolist}. 
The resulting interface of the implemented game is shown in Figure \ref{fig:myfeh_endresult}, with the given interface after selecting a unit in Figure \ref{fig:myfeh_selecting} and initiating a combat in Figure~\ref{fig:myfeh_combat}.

The difference between the two scenes lies in the map layout, data collection, and enemy behavior.

\begin{figure}[b]
    \centering
    \includegraphics[width=\linewidth]{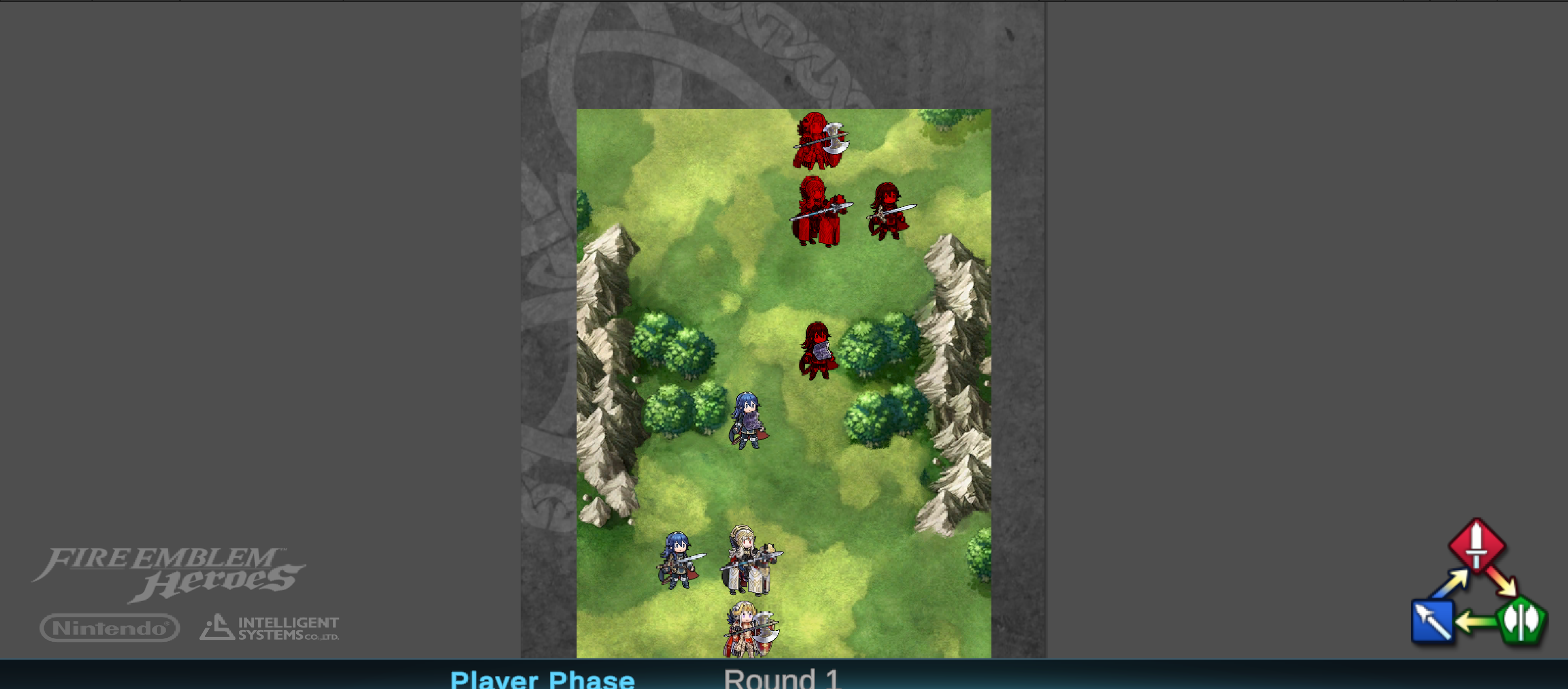}
    \caption{Example of Mirror Mode starting positions.}
    \label{fig:MirrorMode}
\end{figure}

The standard scene functions as a platform for collecting data, as well as training agents. The mirror scene is solely used for playing Mirror Mode, to evaluate the effect of the trained agents to imitate the player's strategy. In this scene, the enemy team is a complete mirror of the player's team, including unit types, weapons, and positions, as shown in Figure \ref{fig:MirrorMode}.

\subsubsection{Standard Enemy Behavior}

\begin{algorithm}
\caption{Standard Mode enemy behavior script.}\label{alg:standard_enemy}

\KwResult{$a, t, u$} %include set of 0,1,2 for actions
$a \in \ A(s)$ \\
$t \in \text{Tiles}_{6\times8}$ \\
$u \in \text{U} = \{0, 1, 2, 3\}$

\BlankLine

\While{$|E| > 0$}
{
    $e \gets SelectUnit(E)$\\
 
    \If{$U_{inrange}$ from $e == 1$  \textbf{and} $T_{inattackrange} \neq \varnothing$}
    {
    $a \gets 2$\\
    $t \gets t \in T_{inattackrange}$\\
    $u \gets o \in U_{inrange}$\\
    \textbf{perform} $e$ attacks $u$ on $t$ \\
    \textbf{return}
    }
    \ElseIf{$U_{inrange}$ from $e >   1 $ \textbf{and} $T_{inattackrange} \neq \varnothing$}
    {
        $a \gets 2$\\
        $t \gets t \in T_{inattackrange}$\\
        $u \gets u\in U_{inrange}, $ where $\max(DamageTo[u, e])$\\

        \textbf{perform} $e$ attacks $u$ on $t$ \\
        \textbf{return}
    }
    \ElseIf{$T_{inrange} \neq \varnothing$}
    {
    $a \gets 1$\\
    $t \gets t \in T_{inrange}$, where $min(dist[t, o])$\\
    $u \gets null$
    \textbf{perform} $e$ moves to $t$
    \textbf{return}
    }
    \Else{
    $a \gets 0$\\
    $t \gets t_{current}$\\
    $u \gets null$ \\
    \textbf{perform} $e$ stays idle \\
    \textbf{return}
    }
    \textbf{remove} $e$ from $E$\\
}
\end{algorithm}

For the Standard Mode, a rule-based enemy AI is implemented based on the derived gameplay observations by expert players, documented by Game8 Inc. \cite{enemyAI}. 
The algorithm follows a greedy approach, summarized in Algorithm \ref{alg:standard_enemy}, where enemies always attack if there is an opposing unit within their attack range. When multiple targets are in range, the enemy chooses the unit to which it can deal the most damage.
If no opposing units are within attack range, the enemy holds its position.
Furthermore, the order in which enemy units take their turns is determined by the following a set of priorities. Melee attackers are prioritized over ranged attackers, and are allowed to attack first. Among units with the same attack range, it prioritizes the units that can reach the player units the fastest. If all factors are equal, the leftmost unit in the map acts first.

\subsubsection{Mirrored Enemy Behavior.}\label{subsec:MirrorMode}

The enemy action decision process in the mirror scene is purely controlled by the learned algorithm, provided by a trained model. In the enemy's turn, the agent requests a decision from the trained model. This returns the discrete actions array based on the collected observation. The enemy agent uses the array to set the actions. A summary of this enemy behavior is provided by Pseudocode \ref{alg:mirror_enemy}.

\begin{algorithm}
\caption{Mirror Mode enemy behavior script.}\label{alg:mirror_enemy}

\KwResult{$a, t, u$} %include set of 0,1,2 for actions
$a \in \ A(s)$ \\
$t \in \text{Tiles}_{6\times8}$ \\
$u \in \text{U} = \{0, 1, 2, 3\}$

\BlankLine

\While{$|E| > 0$}
{
    $e \gets SelectUnit(E)$\\
 
    $A \gets$ RequestDecision(s)\\
    $a \gets A[0]$\\
    $t \gets A[1]$ \\
    $u \gets A[2]$\\

    \If{$a$ == $2$ and $t$ not in attack range or $u$ not in attack range}
    {
        \If{$T_{inattackrange}$}{$t \gets random(T_{inattackrange})$}
        \ElseIf{$t$ in movement range}{$a \gets 1$}
        \Else{$a \gets 0$}
    }
    \ElseIf{$t$ not in movement range}
    {
        $a \gets 0$
    }

    \textbf{perform} action $a$ with tile $t$ and target $u$ for enemy $e$
    \textbf{remove} $e$ from $E$\\
}
\end{algorithm}

\subsubsection{Player Agent Behavior Script.} \label{subsec:agentscript}
The player's actions are handled through the mouse input system. Selecting units, tiles, and targets are all handled through intuitive clicking interaction, based on the original game. Each player unit contained its distinct recording component and agent behavior script, available through the ML-Agents package \cite{ML-Agent}. Once an action is put through, the script processes the observed state and taken action by the player, and stores the corresponding state-action pair in  demonstration files.

\section{Agent}
The approach for the research begins with the underlying programming for training the Mirror Mode AI and collecting data. This section describes the dataset that is used for training the models, including the methods for collecting the data. Followed by our used training methods that incorporates the RL algorithm PPO, and IL approaches BC and GAIL. RL makes use of a reward system whereas the collected dataset is only utilized by the IL algorithms.

% \subsection{Architecture Overview}

% omschrijven wat de uiteindelijke agent heeft gebruikt.
% PPO for RL wat dus de geimplementeerde st en ac space heeft gebruikt. En dan GAIL en BC die de collected data gebruikne
% Dit wordt hier verder uitgelegd.

\subsection{Dataset}
Player demonstrations were collected from real-time games played in the standard scene. 
The scene includes three augmented environments, that were created by flipping the original map configuration along the x-axis, y-axis, and both axes, as presented in Figure \ref{fig:mirroredenvs}.  
This allowed a single set of demonstrations to generate four unique moves at one time step. 
The collected demonstration data is stored in $.yaml$ files saved in the game scene directory.

\subsection{Agent Training}
\begin{figure}[t]
    \centering
    \begin{subfigure}[b]{0.4\textwidth}
        \centering
        \includegraphics[width=\textwidth]{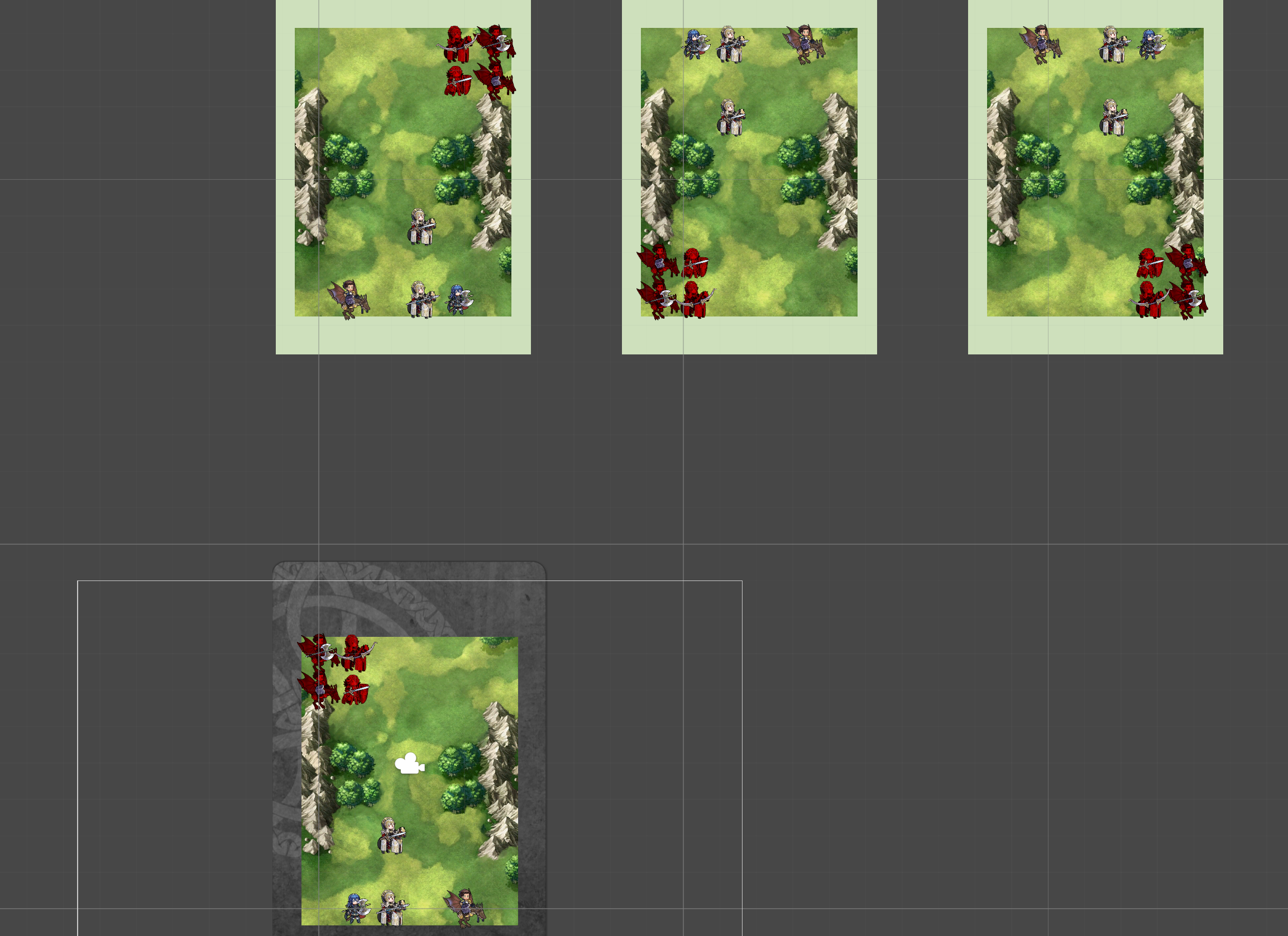}
        \caption{Augmented environments presented above the one original environment, for faster data collection.}
        \label{fig:mirroredenvs}
    \end{subfigure}
    \hfill
    \begin{subfigure}[b]{0.4\textwidth}
        \centering
        \includegraphics[width=\textwidth]{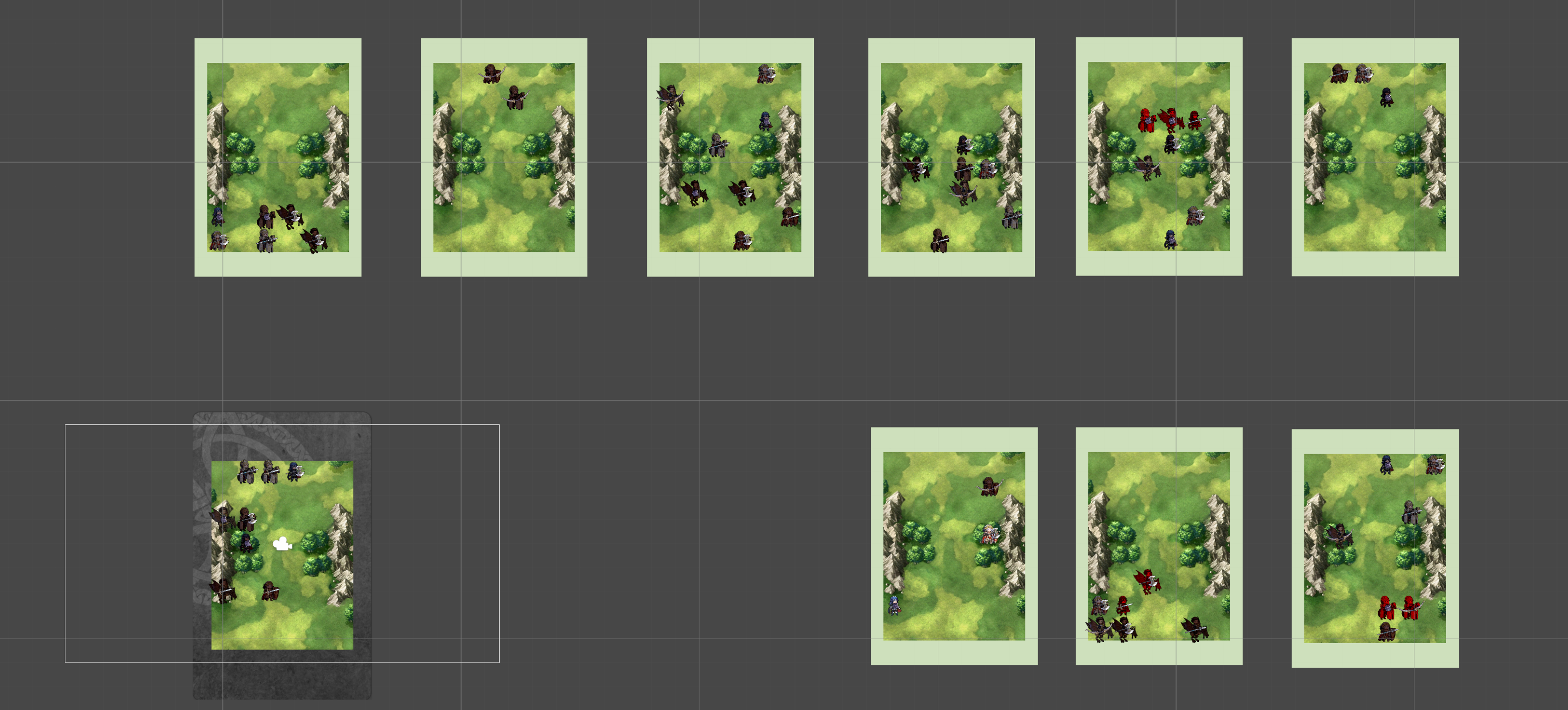}
        \caption{Environment copies running in parallel for model training, each controlling and playing their own game.}
        \label{fig:training_scene}
    \end{subfigure}

    \caption{Environmental setup for collecting data and training agents.}

\end{figure}

Training agent models for the Mirror Mode enemies occurred in the standard scene as well. Alongside the four augmented environments, six more environments were created to accelerate training. In total, all ten environments functioned as direct copies from the original environment and operate in parallel only when training was enabled (Figure \ref{fig:training_scene}). 

During training, the agent scripts in Unity were activated, that use the directed information available in a \textit{.yaml} format. This file includes the demonstration path from the collected player data, and the desired parameters and algorithm names for the model.
The unit behaviors script handled the interaction input explained in \ref{subsec:agentscript}, as well as the functions enabling RL. It first collects observations from the current state of the map based on the agent's local position. Then, invalid actions are masked from the model, preventing these actions from getting selected. For this study, the function filters out the tiles that are not reachable for an agent at a current state, and the targets that are not available to attack. Lastly, it receives the selected action given through the mouse interaction provided by the user.
% To train the model based on IL algorithms, the demonstration folder directory must be specified in the \textit{demo\_path} in the configuration file that is used for training.
% To start training, the virtual environment described in section \ref{sec:setup} needs to be activated through the command prompt. Training can be activated when using the following command: \newline $\textbf{mlagents-learn } \textit{directory\\filename.yaml --  run-id=ID-Name-Run}$.
% \newline
Once the game scene is started in Unity, the training progression begins.

\subsection{Reinforcement Learning}
The used RL algorithm, PPO, requires a defined action space, state space, and reward system. This section presents how the three were implemented for this study.

\subsubsection{Action Space}

The actions space is given by an array of three discrete action variables: action type, selected tile, targeted unit. 

An action type holds an index that corresponds to the action chosen by the player, which can be to wait (0), to move (1), or to attack (2).

The selected tile corresponds to the index of a tile within the map size \textit{6x8}, resulting in an array of size 48. This tile index then maps to the tile that the unit needs to move to. For the waiting action, this equals the index of the tile that the unit is currently already on. For the attack action, the tile that the unit stands on to launch it attack from is used. 
    
Lastly, the targeted unit decides the target that will be attacked during an initiated combat. If no combat is initiated, the value is of no use. In case an attack is performed, the index of the target unit is used. 

\subsubsection{State Space}

The state space of the game is constructed by an array with size 136. This array stores all relevant information concerning the unit's and enemy's stats, weapon, and unit types, and the manhattan tile distance between the unit and an opponent are stored in the array. For more compatible computation all values are normalized to a value between 0-1.

\subsubsection{Reward System}

Finding a proper reward system for the game environment acquired testing and time. A reward range of \textit{[-1,1]} was followed, recommended by the package creators \cite{ML-Agent-Reward}.
Killing enemies and winning a round was rewarded by a positive reward of one. A negative reward of one was given to a unit after it had died, or when the game round was lost. Any other rewards for choosing valid actions received a reward of +0.3. 
To stimulate training, a maximum set of actions per game was implemented and set to a value of 20. After reaching the maximum value, the game ends in a tie resulting in a punishment of -1. This improved the learning curve, making sure the agents did not avoid attacking targets. 

Finding these settings, the agents were fully ready to learn from the environment and demonstrations, with no further complications that led to unjustified behavior. The rest of the research continued with these agent settings.

% \begin{figure}[t]
%     \centering
%     \includegraphics[width=0.5\linewidth]{Images/stuck_corner.png}
%     \caption{Early stage of the developed strategy game. The agents found a tactic to move to the lower left corner and stay there to protect themselves.}
%     \label{fig:agents-stuck}
% \end{figure}

%image stuck corner
% Configurations are tracked using tensorboard logdir. It can be loaded on the localhost port using the following command: tensorboard --logdir results
% For more insights, custom metrics are added. 

\subsection{Imitation Learning}
For the IL behavior of the agent, we employ BC and GAIL algorithms developed by Juliani et al. \cite{juliani2020}. IL is used in conjunction with RL, to enable better copying behavior and reduce the time it takes for an agent to acquire a proper behavior.
The collected dataset provides the expert demonstrations from which the IL algorithms lean a policy. 
GAIL rewards the agent using an adversarial approach that evaluates how closely its actions match those in the demonstrations.
%Training this model updates a discriminator that learns to distinguish agent moves and demonstration moves. The discriminator returns a reward based on its certainty that the action was provided by the demonstrations, which the agent tries to maximize by taking actions as similar as possible to the demonstrations. 
The GAIL loss serves as a penalty for the discriminator when it misclassifies an agent action as a demonstration action or vice versa. We aim to keep this loss at a moderate level and use it as an indicator of how well the agent imitates the demonstrations.
BC is used to pre-train the agents before applying RL and GAIL. It updates the agent's policy to exactly replicate the actions the demonstration set, and is therefore most effective when a sufficiently large set of demonstrations is available. Consequently, BC is used in combination with GAIL and RL to leverage imitation behavior. 

The degree to which the agent relies on demonstration data can be set manually through the strength parameter. For both BC and GAIL, various values are tested in the firs half of experiments, to provide an optimal configuration. 

\section{Results} \label{sec:modeltesting}
A set of optimal hyperparameters and algorithm combinations to train the Mirror Mode model, were found through the first two conducted experiments. These two experiments form the first half of the experimental setup for this research paper, to find an answer to the question whether RL and IL can be used to imitate player's strategy in video games. This is followed by the third experiment, involving user studies to evaluate the effectiveness of the agent models on game experience and its imitation capabilities.

\subsection{Experiment 1: Hyperparameter Optimization}\label{exp:modelfinetuning}
\begin{table*}[tbp]
\caption{Hyperparameter tuning overview. The diagonals show the values for the hyperparameters that are tested. Other values are fixed and are used for testing the hyperparameter in the corresponding row.}
\label{tab:tuning_matrix}
\begin{tabular}{|p{0.15\textwidth}|p{0.15\textwidth}|p{0.15\textwidth}|p{0.15\textwidth}|p{0.15\textwidth}|p{0.15\textwidth}|}
\toprule
Tuned & PPO $\alpha$ & GAIL $\alpha$ & BC str & Curiosity str & Extrinsic str \\
\midrule
\textbf{PPO} $\alpha$
& \textbf{0.0005, 0.0003, 0.0001} 
& 0.0001 
& 0.0 
& 0.0 
& 0.0 \\

\hline

\textbf{GAIL} $\alpha$
& 0.0003 
& \textbf{0.001, 0.0005, 0.0003, 0.0001} %0.0001=2206, 0.001=2206_3, 0.0005=2206_4, 0.0003=2206_5
& 0.0 
& 0.0 
& 0.0 \\

\hline

\textbf{BC str} 
& 0.0003 
& 0.0001 
& \textbf{0.4, 0.5, 0.6, 0.8, 1.0} %0.5=2406, 0.6=2406_2, 0.8=2406_3, 1.0=2406_4, 0.4 = 2406_5
& 0.0 
& 0.0 \\

\hline

\textbf{Curiosity str} 
& 0.0003 
& 0.0001 
& 0.4
& \textbf{0.0, 0.05, 0.1} %0.0= 2506, 0.02=2506_2--> niet afgemaakt, 0.05=2506_3, 0.1=2506_4
& 0.0 \\

\hline

\textbf{Extrinsic str} 
& 0.0003 
& 0.0001 
& 0.4 
& 0.1 
& \textbf{0.1, 0.5, 1.0} \\ %extrinsic reward was added in 2506_5, had a visible effect to cumu reward, whereas curiosity had little effect. For optimal imitation behavior it is chosen to only use extrinsic reward and no curiosity
\hline
\end{tabular}

\end{table*}

The aim of the first experiment was to optimize the performance and imitation quality of the agents, by adjusting one parameter at the time, while keeping the others constant. The value range of each parameter recommended by the ML-Agent developers were taken into consideration \cite{unity-ml-agents-parameters}.

Optimizing the models for this study purely focused on the following hyperparameters:
\begin{itemize}
    \item BC strength;
    \item PPO learning rate $\alpha$;
    \item GAIL learning rate $\alpha$; 
    \item extrinsic strength;
    \item curiosity strength.
\end{itemize}

For each tested hyperparameter, the used values for testing are summarized in Table \ref{tab:tuning_matrix}. The rows show the parameters that are tuned, and the columns show the values set to the parameters while tuning. The tested values for each parameter are presented in bold font. 
Remaining parameters not mentioned in the list are set to their default values.
Each model was trained for 200,000 steps starting at step 0 with no prior knowledge yet. Models were evaluated based on the total cumulative reward and GAIL loss. Cumulative reward served as a measurement for learning capability of the agents, and GAIL loss for the ability of imitating a player's strategy.

% \begin{table*}
%   \caption{Some Typical Commands}
%   \label{tab:commands}
%   \begin{tabular}{ccl}
%     \toprule
%     Command &A Number & Comments\\
%     \midrule
%     \texttt{{\char'134}author} & 100& Author \\
%     \texttt{{\char'134}table}& 300 & For tables\\
%     \texttt{{\char'134}table*}& 400& For wider tables\\
%     \bottomrule
%   \end{tabular}
% \end{table*}

% For identifying the best learning rate for PPO and GAIL, and the best strength for BC, no extrinsic rewards or curiosity parameters were used. The goal was to find suitable values for these parameters to stimulate learning by keeping track of the cumulative reward over time, but also to encourage imitating player demonstrations by maintaining a proper GAIL discriminator loss around a value of 0.5. 

\subsection{Experiment 2: Model Configurations} \label{exp:modelconfigurations}
After identifying the optimal performing model from Experiment \ref{exp:modelfinetuning}, further experiments were conducted to determine the optimal configuration for the enemy AI in Mirror Mode. Several combinations of RL and IL techniques were tested. 

The following model variants were evaluated:
\begin{itemize}
    \item PPO only %-->2606\_4;
    \item PPO + GAIL %--> 2606\_3;
    \item PPO + GAIL + BC % 2406_5; ppogailbc
    \item PPO + GAIL + BC + Curiosity; %2506_4
    \item PPO + GAIL + BC + Curiosity + Extrinsic Rewards; %2506_6
    \item PPO + GAIL + BC + Extrinsic Rewards; %2606_2
    \item PPO + GAIL + BC + Extrinsic Rewards + self-play %--> selfplay
\end{itemize}

% compare combinations: ppo (2606_4), ppo+gail (2606_3), ppo+gail+bc (ppogailbc), pp+gail+bc+cur (2506_4), ppo+gail+bc+cur+extr (2506_6), ppo+gail+bc+extr (2606_2)

%compare variant training styles: selfplay2 (selfplayTest4), final model with new batch size etc (FinalExperimentalModel), model with optimal pars from previous tests (2706_2), weapon triangle (2605_5)

%2605_5 also tests purely weapon triangle, to reduce dimension of attack learning space to possible copy behavior better.

\begin{table}[t]
\caption{Hyperparameters used during model combination testing.}
\label{tab:parameters_modelcombination}

\begin{tabular}{lc}
\toprule
Parameter & Value \\
\midrule
    PPO learning rate & 0.0003 \\
    PPO hidden units & 256 \\
    PPO batch size & 128 \\
    GAIL learning rate & 0.0001 \\
    GAIL hidden units & 64 \\
    GAIL gamma & 0.85 \\
    GAIL strength & 1.0 \\
    BC strength & 0.5 \\
    Extrinsic strength & 0.9\\
\bottomrule
\end{tabular}
\end{table}

Similar to the first experiment, the cumulative reward and GAIL loss served as performance evaluation metrics, taken over 200,000 steps.

During this phase, a fixed set of hyperparameters was used for consistency across models, as listed in Table \ref{tab:parameters_modelcombination}. 

To increase imitation quality, a final configuration is tested with increased values for the GAIL hidden unit and PPO batch size. Based on the results of the model combination experiment, a model is chosen to see the effect of a GAIL hidden unit value of 128, together with a PPO batch size of 256.

\subsection{Reflection Model Results} \label{subsec:resultsmodeltesting}
% The first part of this study looked into the possible hyperparameters and configurations for the model that will be applied to the Mirror Mode agents. Several tests were conducted to find a model combination most suitable for the agent behavior to imitate player strategies, as previously discussed in subsections \ref{exp:modelfinetuning} and \ref{exp:modelconfigurations}. 
The results of the two conducted model configuration experiments are discussed in this subsection.

\subsubsection{Model Finetuning}
% The first experiment tested several hyperparameters from different RL and IL techniques, mentioned in Table \ref{tab:tuning_matrix}. 

\begin{figure*}[tbp]
    \centering

    % Row 1
    \begin{subfigure}{0.48\linewidth}
        \centering
        \includegraphics[width=\linewidth]{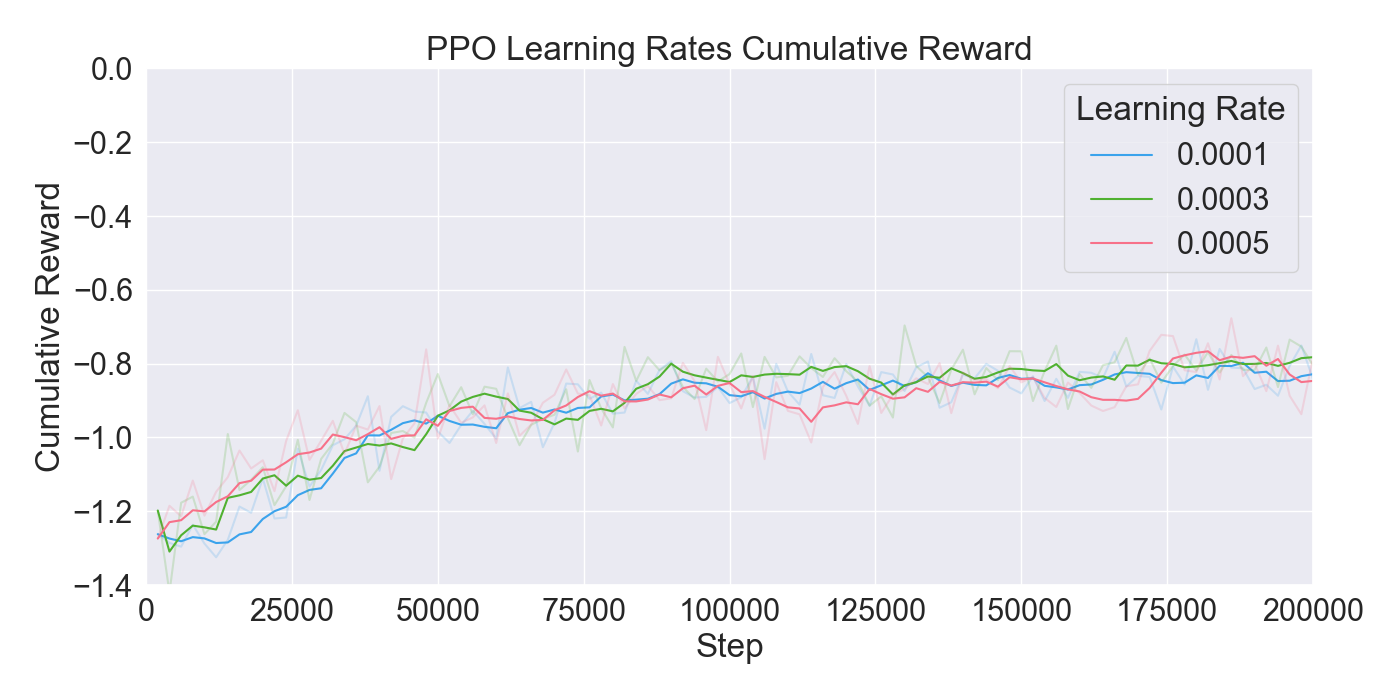}
        %\caption{Model tested on PPO learning rate.}
        \label{fig:FineTuning_PPO_a}
    \end{subfigure}
    \hfill
    \begin{subfigure}{0.48\linewidth}
        \centering
        \includegraphics[width=\linewidth]{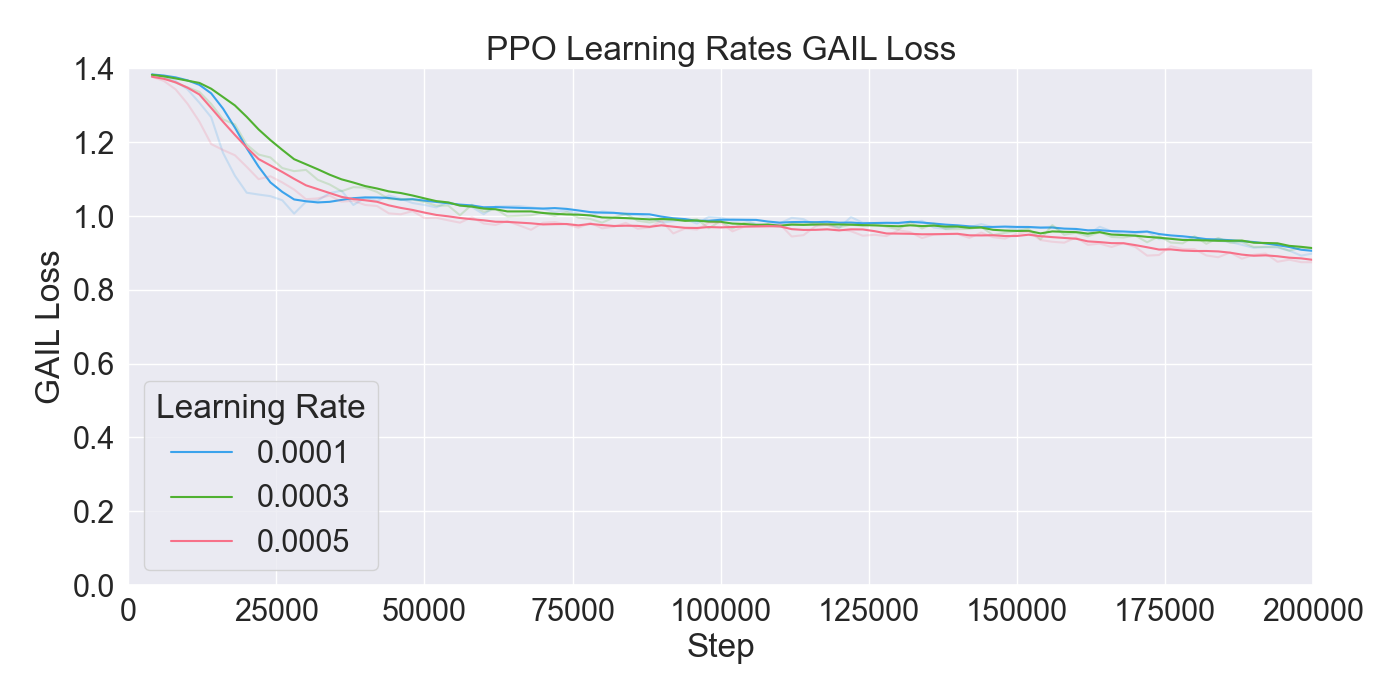}
        %\caption{Model imitation quality.}
        \label{fig:FineTuning_PPO_b}
    \end{subfigure}

    \vspace{0.2em}

    % Row 2
    \begin{subfigure}[b]{0.48\linewidth}
        \centering
        \includegraphics[width=\linewidth]{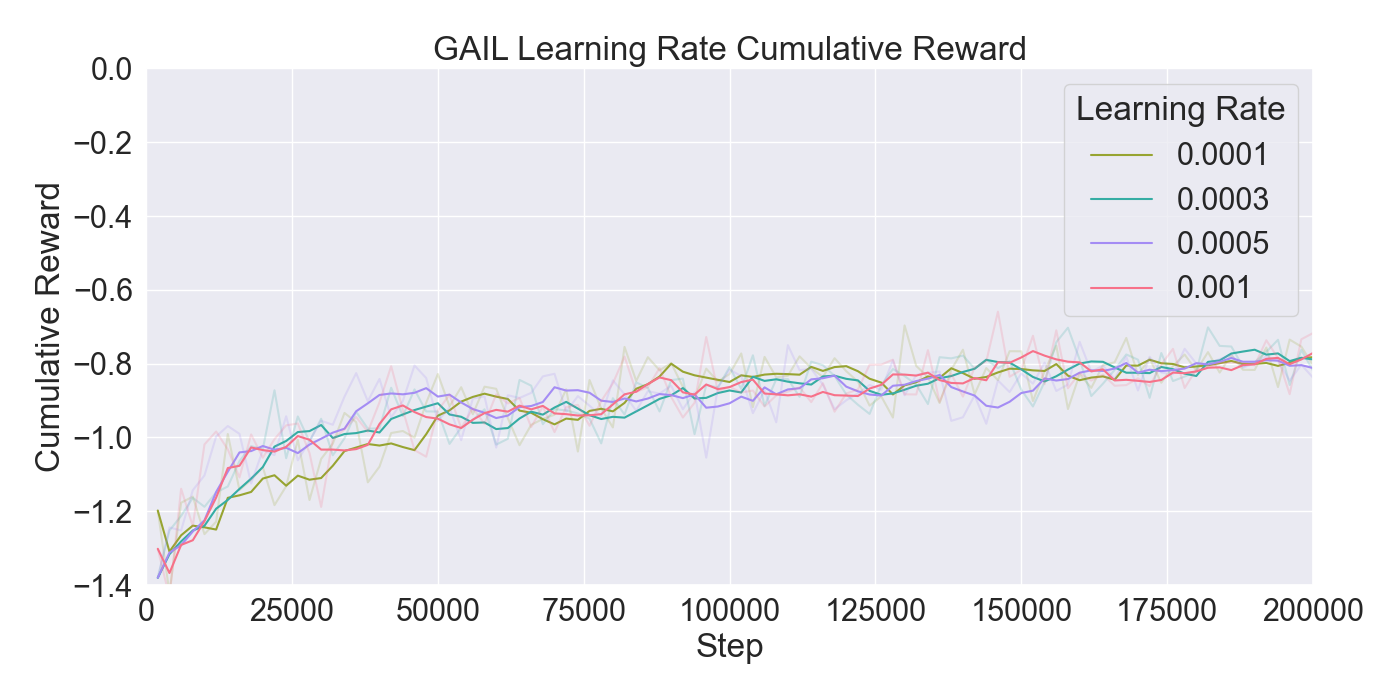}
        %\caption{Effect of GAIL $\alpha$ on agent's performance in FE enviroment.}
        \label{fig:FineTuning_GAIL_a}
    \end{subfigure}
    \hfill
    \begin{subfigure}[b]{0.48\linewidth}
        \centering
        \includegraphics[width=\linewidth]{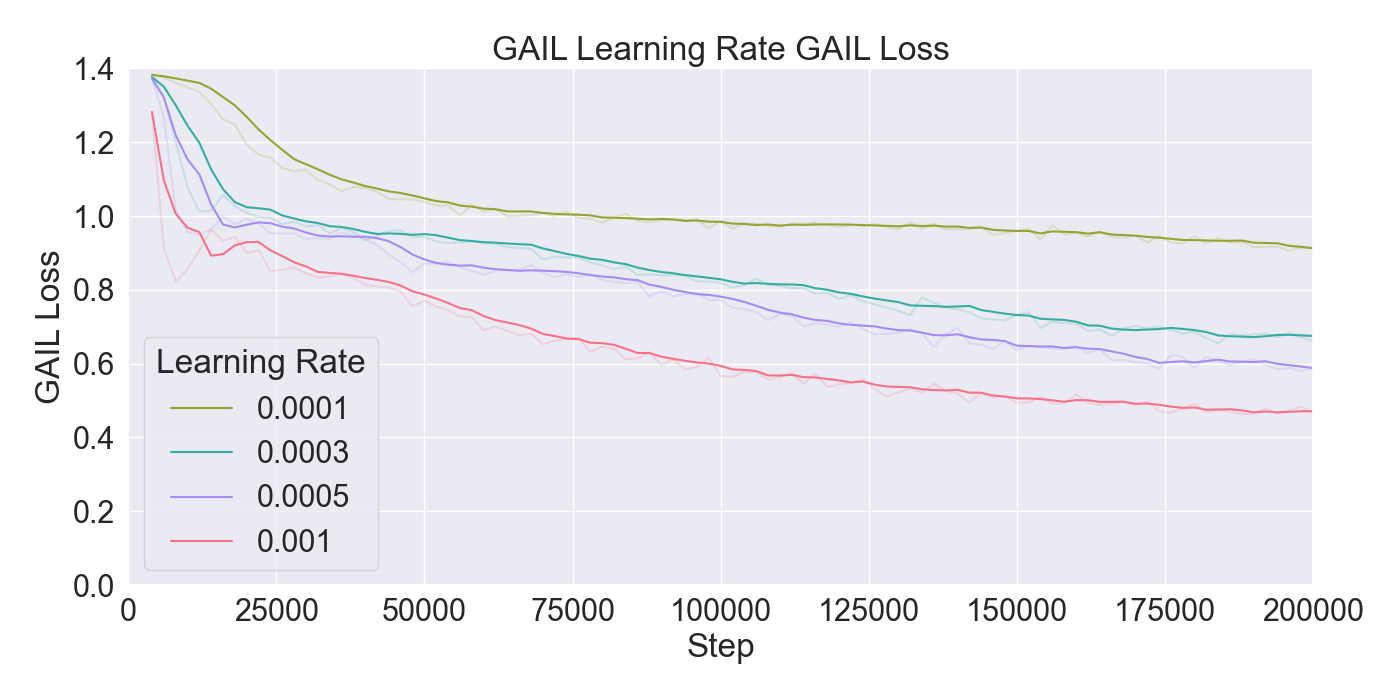}
        %\caption{Effect of GAIL $\alpha$ on agent's imitation quality of the player's demonstrations.}
        \label{fig:FineTuning_GAIL_b}
    \end{subfigure}

    \vspace{0.2em}

    % Row 3
    \begin{subfigure}[b]{0.48\linewidth}
        \centering
        \includegraphics[width=\linewidth]{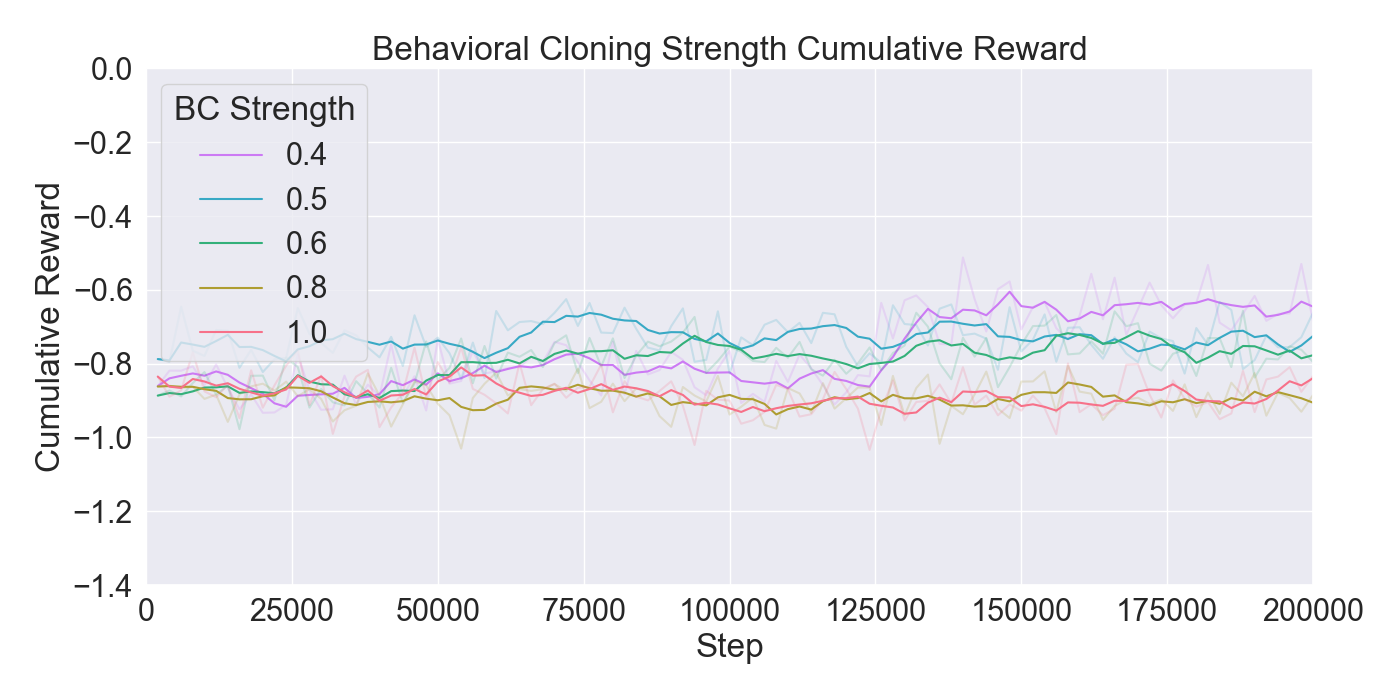}
        %\caption{Effect of BC strength on the agent's performance in FE environment.}
    \end{subfigure}
    \hfill
    \begin{subfigure}[b]{0.48\textwidth}
        \centering
        \includegraphics[width=\textwidth]{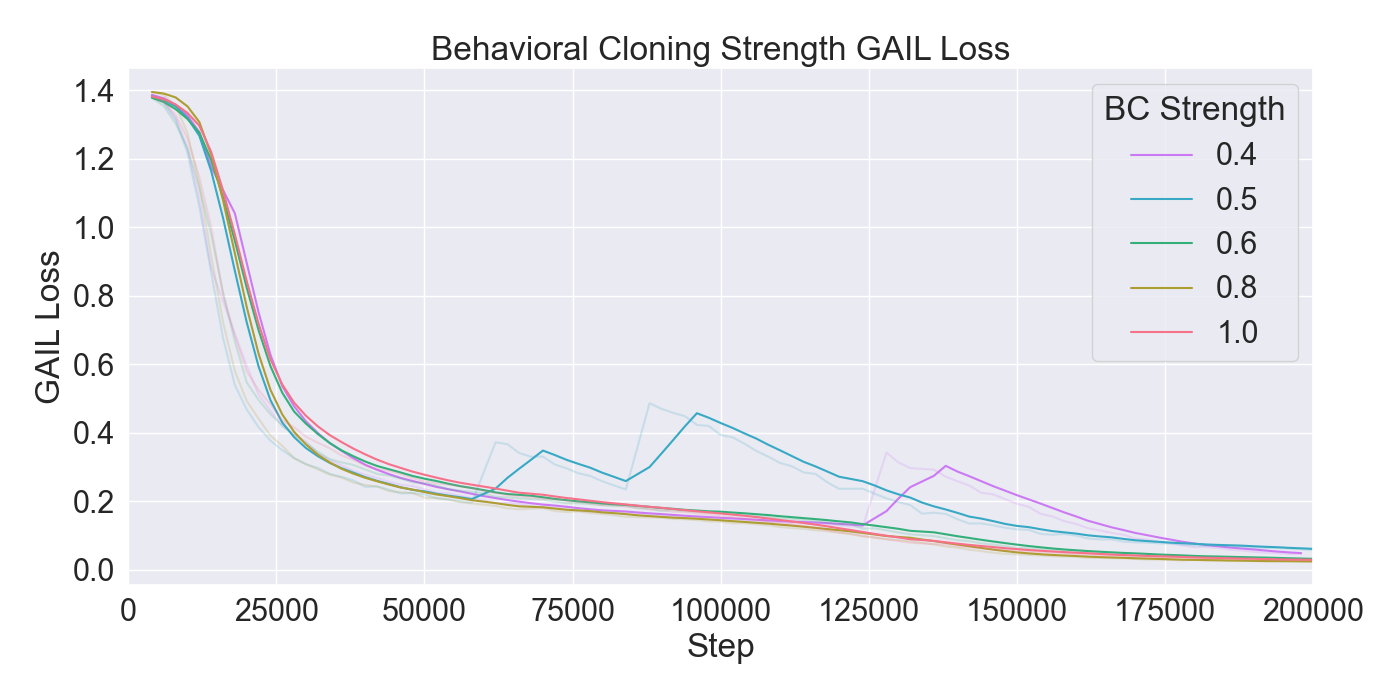}
        %\caption{Effect of BC strength on agent's imitation quality of player demonstration.}
    \end{subfigure}

    \vspace{0.2em}

        % Row 1
    \begin{subfigure}[b]{0.48\textwidth}
        \centering
        \includegraphics[width=\textwidth]{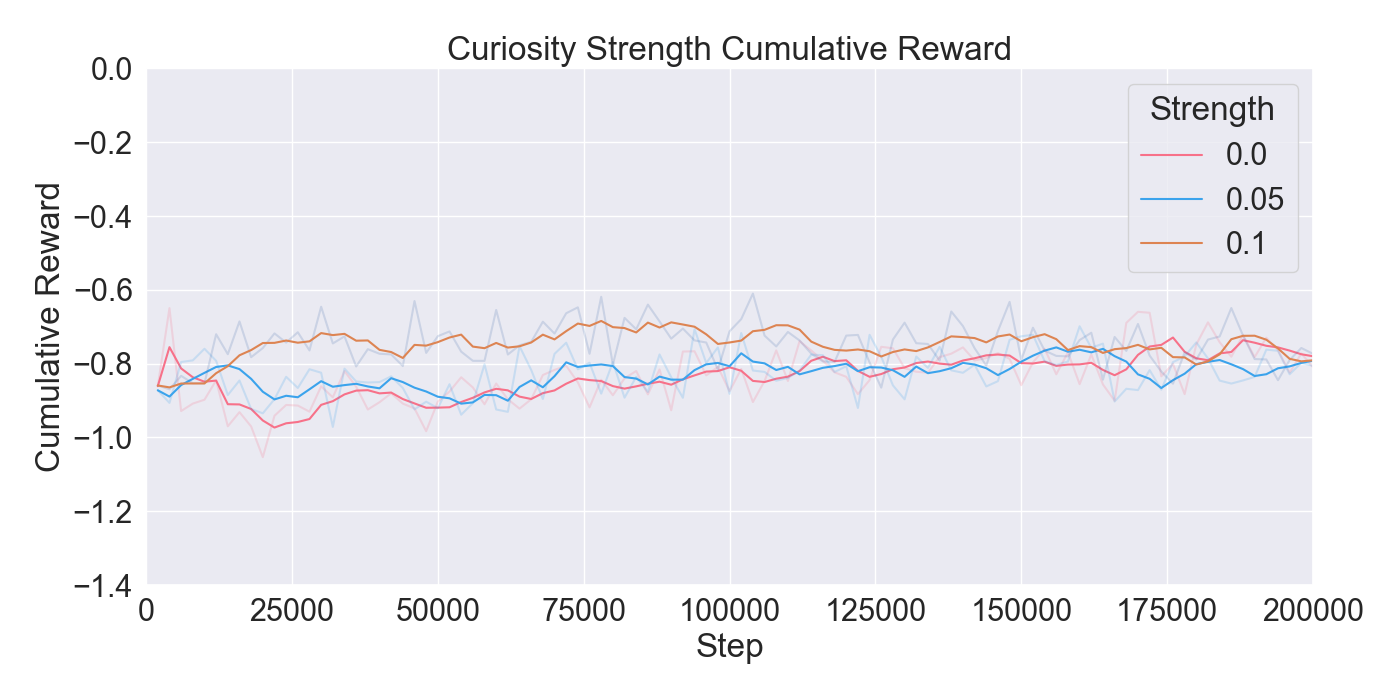}
        %\caption{The cumulative reward for different curiosity strength values. Curiosity 0.1 seemingly outperforms the other values. All values converge to a similar cumulative reward level, while a strength set to 0.1 appears to be give a slightly higher mean reward.}
    \end{subfigure}
    \hfill
    \begin{subfigure}[b]{0.48\textwidth}
        \centering
        \includegraphics[width=\textwidth]{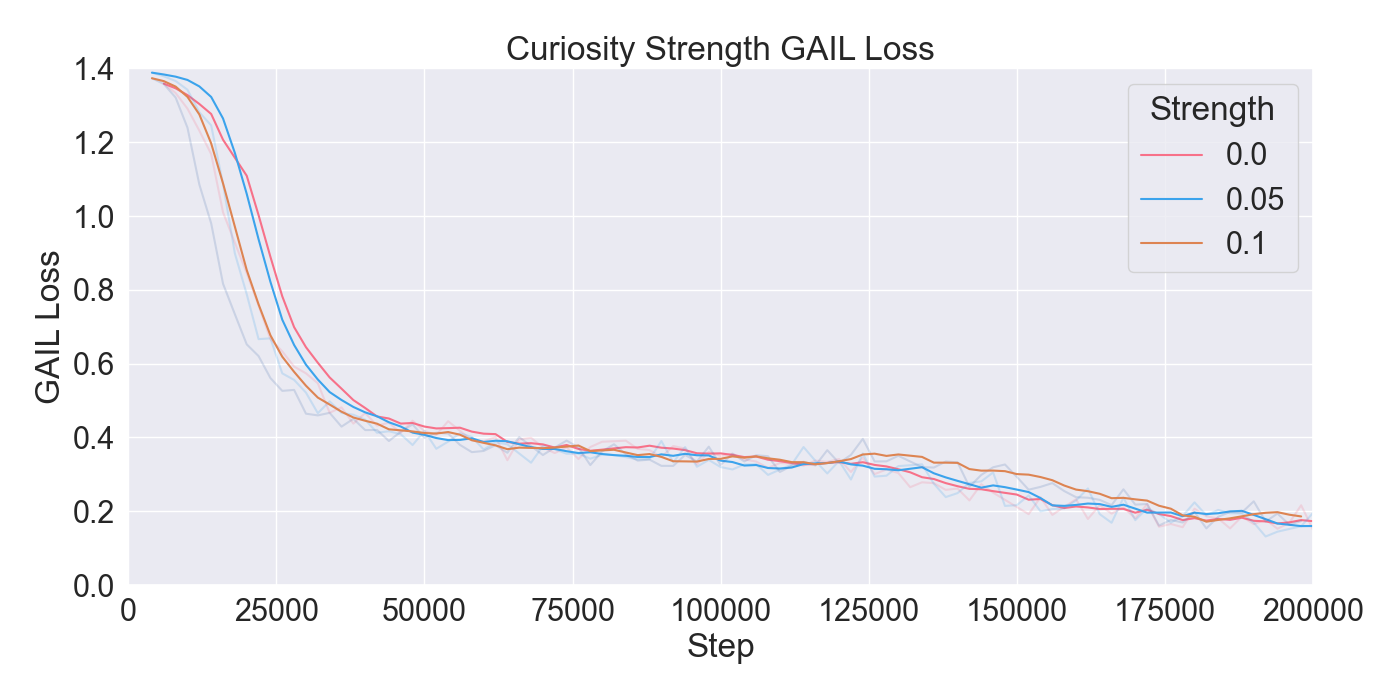}
        %\caption{The GAIL discriminator loss set for several curiosity strengths. There seems to be little difference in loss when curiosity is applied, compared to when the curiosity reward is disabled. Overall, adding curiosity seems to drop the loss more rapidly.}
    \end{subfigure}

    \vspace{0.2em}

    % Row 2
   \begin{subfigure}[b]{0.48\textwidth}
        \centering
        \includegraphics[width=\textwidth]{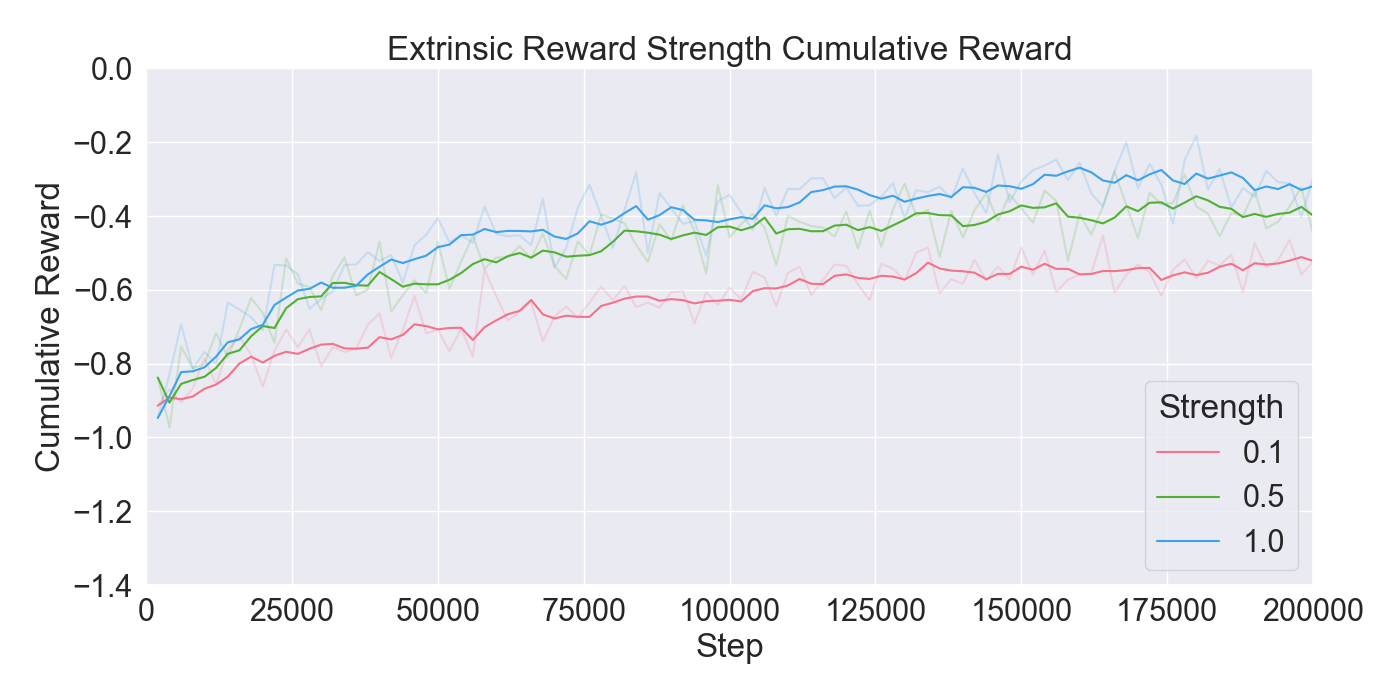}
        %\caption{Cumulative reward given by different extrinsic strength configurations. An immediate improvement in learning can be observed, with strength=1.0 achieving the highest reward. Thus extrinsic rewards highly benefit learning abilities.}
    \end{subfigure}
    \hfill
    \begin{subfigure}[b]{0.48\textwidth}
        \centering
        \includegraphics[width=\textwidth]{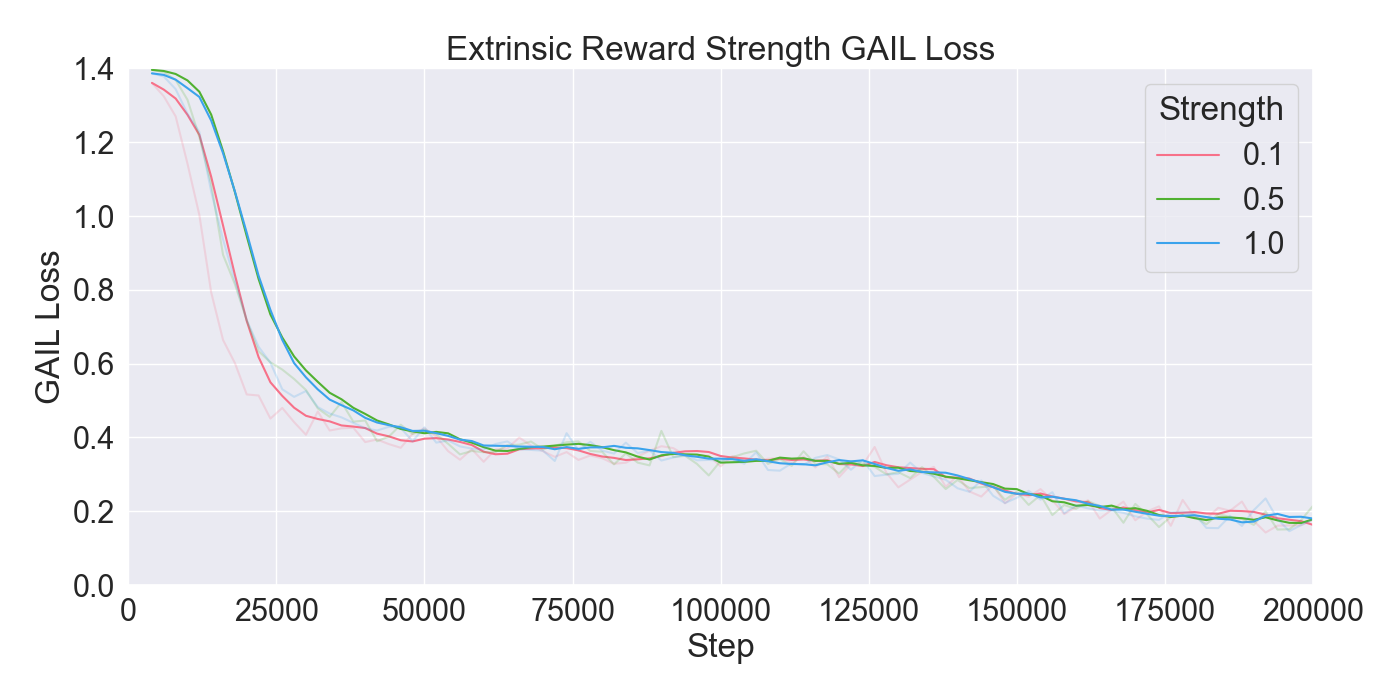}
        %\caption{GAIL discriminator loss set for extrinsic reward strengths. The loss decreases rapidly, with strength=0.1 declining more quickly than the others. Showing that extrinsic rewards possibly negatively impacts agents imitation abilities.}
    \end{subfigure}

    \caption{Hyperparameter finetuning results. Left side presents agent's performance in the FE environment, and right side the imitation quality of player's demonstrations.}
    \label{fig:finetuning_models}
\end{figure*}

The finetuning experiment lead to the results presented in Figure \ref{fig:finetuning_models}. 
The results indicate a gradual learning progression for both PPO and GAIL, whereas BC shows a flattened learning curve. PPO and GAIL converge toward a relatively high cumulative reward of approximately -0.8, with GAIL demonstrating a faster convergence. %For learning rates $\alpha=0.0003$, $\alpha=0.0005$ and $\alpha=0.001$, GAIL reaches a cumulative reward of -1.0 after roughly 25k steps, while PPO achieves this milestone after roughly 40k steps. 
The absence of environment interaction in BC may explain the flat learning trend. Despite the limited learning progression, the BC model achieves a rather good cumulative reward compared to the PPO and GAIL finetuning models. Specifically, a BC strength of 0.4 starts below -0.8, and improves to a higher reward of nearly -0.6, ultimately outperforming the PPO and GAIL.
PPO results in a much higher GAIL discriminator loss, of roughly 0.9, compared to other models, suggesting a poor performance of the GAIL discriminator. Introducing a higher value for GAIL $\alpha$ causes the discriminator loss to drop, showing a much wider range in discriminator loss over the training steps for GAIL compared to other finetuning models. %However, maintaining a balanced discriminator loss is crucial, as excessively low values can cause gradient vanishing, hindering imitation learning. BC further reduces the loss, often pushing it too close to zero. Higher BC strength values amplify this effect, causing a more drastic decrease in the loss.

Interestingly, introducing an extrinsic reward accelerates the learning curve drastically, with the cumulative reward converging to nearly -0.2 when the strength parameter is set to 1.0. However, extrinsic reward reduces the GAIL discriminator loss, dropping too close to zero, which may indicate poor imitation quality. 
In contrast, incorporating a curiosity-based reward appears to have little to no impact on either the cumulative reward or GAIL loss. The results closely resemble those from BC finetuning models without curiosity, suggesting limited effectiveness of curiosity in this setup.

%overall conclusion
%Overall, the results lead to a clear indication of learning capabilities of the agents. A trade-off between imitation ability and cumulative reward performance needs to be maintained. Extrinsic rewards positively affect the cumulative reward, leading to better performing agents. However, dropping the discriminator loss for GAIL, suggesting less suitable imitation behavior. BC and GAIL provide better imitation behavior, but perform less well as they rely less on interacting with the environment. 
Considering these results, it was chosen to continue with a PPO learning rate set to $\alpha=0.0003$, and GAIL learning rate to $\alpha=0.0001$. Moreover, BC was retained to enhance the imitation learning, with its strength set to $0.4$. Curiosity was excluded due to its minimal impact, while extrinsic reward was set to a moderate value of $0.5$ to maintain the balance between imitation and exploration.

\subsubsection{Model Configurations}

\begin{figure}[t]
    \centering
    \begin{subfigure}[b]{0.48\textwidth}
        \centering
        \includegraphics[width=\textwidth]{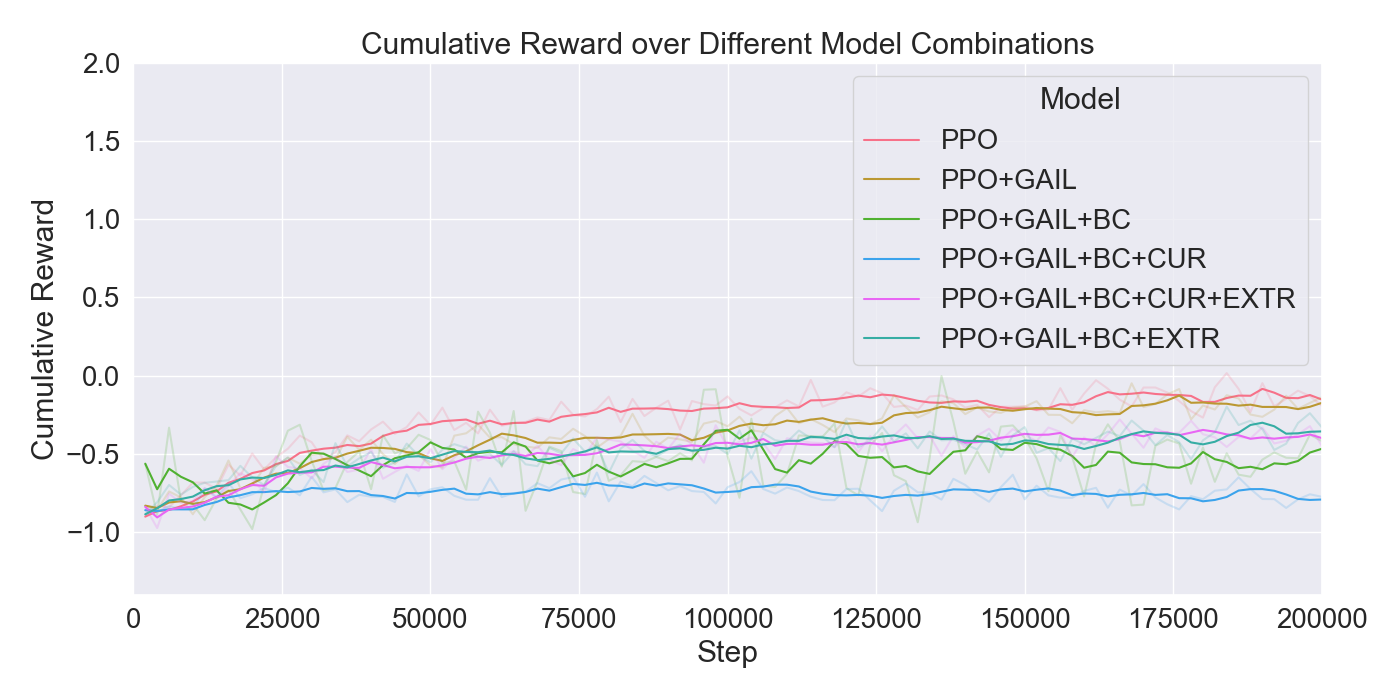}
        %\caption{Cumulative Rewards for different ML combinations. }
        \label{fig:model_combinations_cumureward}
    \end{subfigure}
    \vspace{0.2em}
    \begin{subfigure}[b]{0.48\textwidth}
        \centering
        \includegraphics[width=\textwidth]{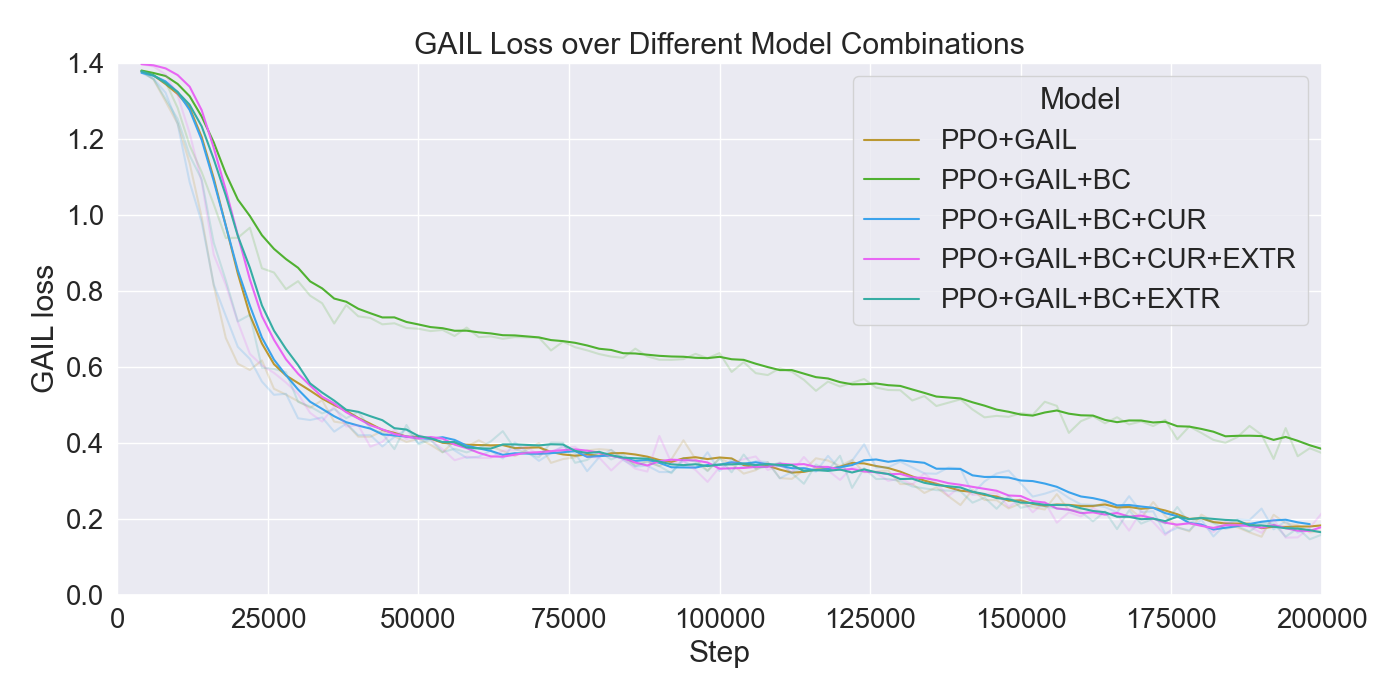}
        %\caption{GAIL discriminator loss for different model configurations.  }
        \label{fig:model_combinations_gailloss}
    \end{subfigure}
    \caption{The results for different RL and IL combinations, over 200k steps. Each model is tested with the parameters found in the experiment \ref{exp:modelfinetuning}.}
    \label{fig:model_combinations}
\end{figure}

\begin{figure}[t]
    \centering
    \begin{subfigure}[b]{0.48\textwidth}
        \centering
        \includegraphics[width=\textwidth]{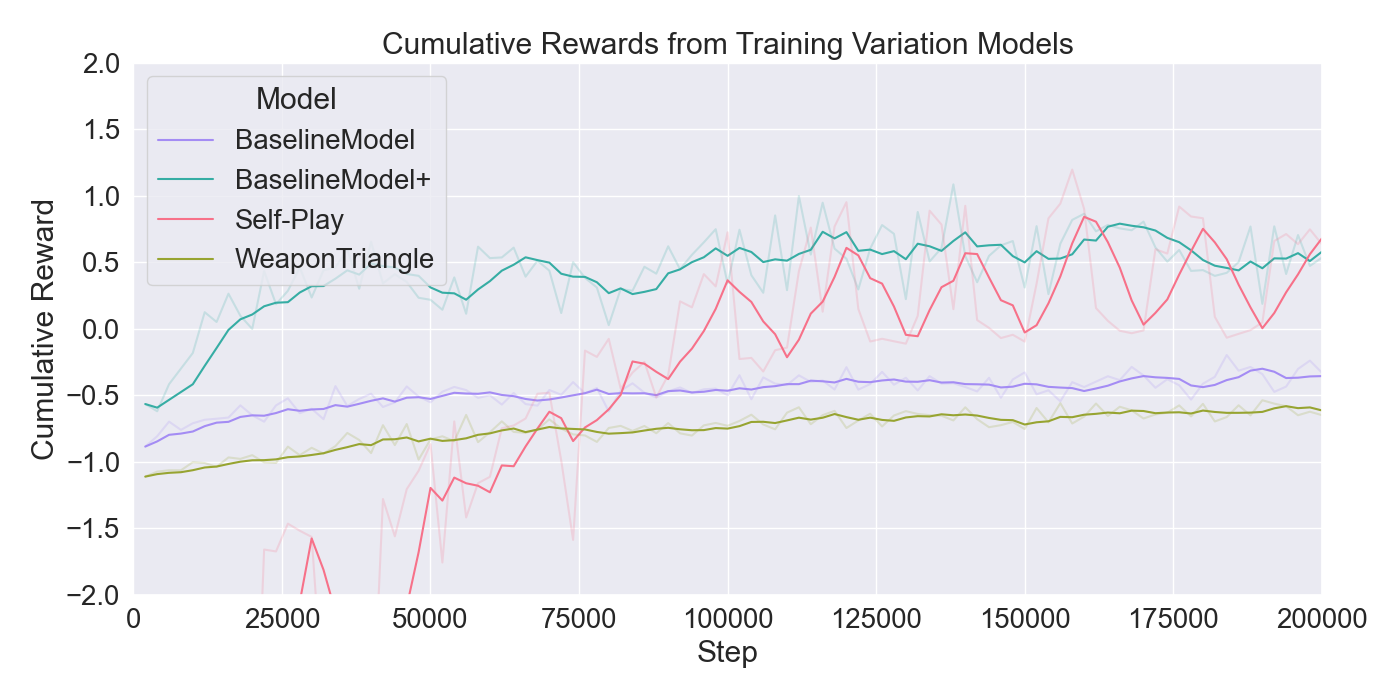}
        %\caption{}
    \end{subfigure}
    \hfill
    \begin{subfigure}[b]{0.48\textwidth}
        \centering
        \includegraphics[width=\textwidth]{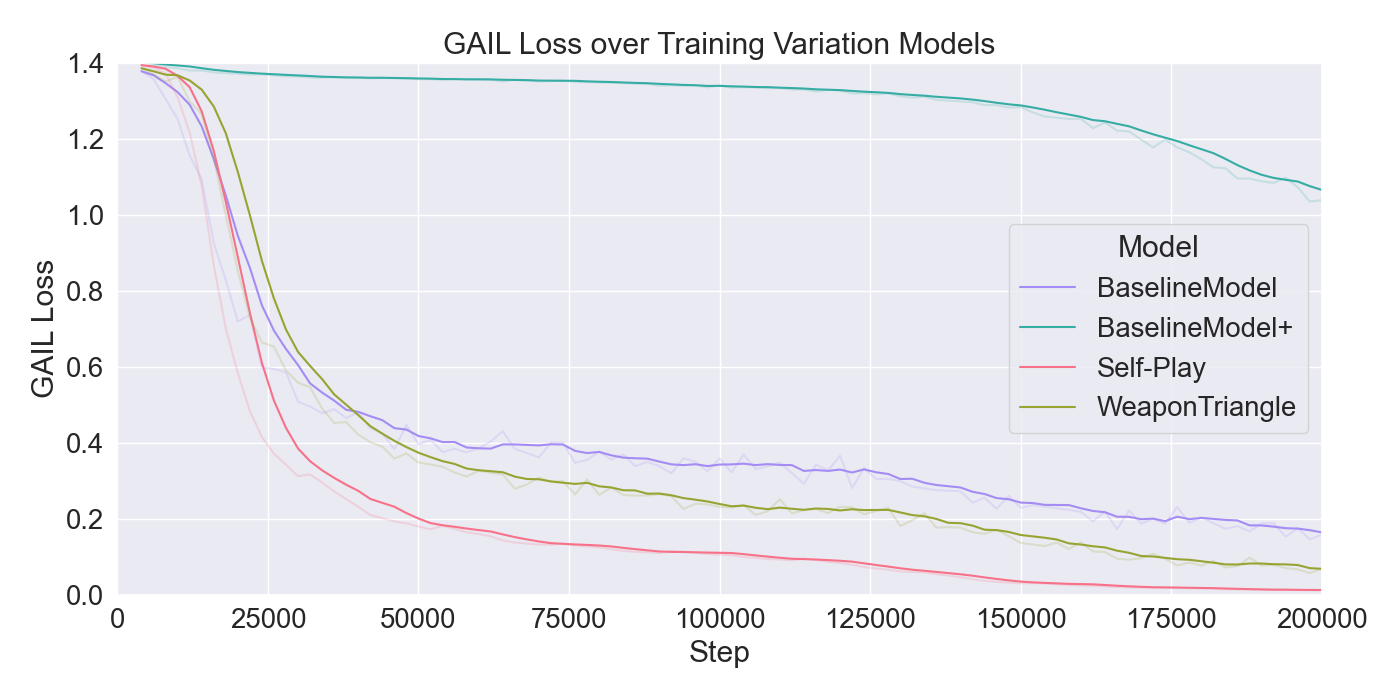}
        %\caption{ }
    \end{subfigure}
    \caption{PPO+GAIL+BC+Extr as BaselineModel performance and imitation behavior compared to alternative models: adjusted hidden units and batchsize in BaselineModel+, training through self-play, and only melee weapons used.}
    \label{fig:training_variants}
\end{figure}

%The second experiment involved different variations of RL and IL techniques to find a suitable model for imitating player strategies. The different combinations are assessed on cumulative reward for overall performance, and GAIL loss as an indicator for imitation performance. 
The results of the model configuration tests are presented in Figure \ref{fig:model_combinations}.
It is observable that the PPO model inclines most rapidly, reaching the highest cumulative reward. On the other hand, the model shows no imitation behavior, as it does not use imitation learning.
$PPO+GAIL$ gradually inclines to a similar performance, but scores a rather low GAIL loss.
The GAIL loss for most models decrease rapidly toward a value of 0.2, except $PPO+GAIL+BC$ which declines more gradually. For a longer training period, it is expected to lose imitation quality as the curves have not reached the minimum.
In this case, the results indicate best imitation practice without extrinsic rewards or curiosity, but a better performance when extrinsic rewards are added to a model using BC. To focus on the imitation abilities, it was chosen to continue with BC, but including extrinsic rewards to allow better performance. To increase the model performance and steadier imitation, the last configuration tests set GAIL hidden units used to 128, and PPO batch size to 256. Additionally, Self-Play is applied to the setup with the found optimal hyperparameters, and another model is trained where the attacks are limited to melee weapons, reducing the action
space. The corresponding results are presented in Figure \ref{fig:training_variants}.

BaselineModel+ converges quickly to a cumulative reward of nearly 0.6 despite a less stable curve. 
Overall, the BaselineModel+ demonstrates strong potential agent behavior.
The GAIL loss shows a slower decrease over time, making it more suitable for training the Mirror Mode models over a larger period of steps.
These results indicate that the modified batch size and hidden units result in a faster improvement in cumulative reward but a slower convergence of the GAIL loss. The Self-Play model shows a promising growth in cumulative reward, going from a large negative value to positive reward. However, unlike the BaselineModel+, Self-Play quickly decreases in GAIL loss, reaching a value too close to zero, making it not a reliable model for imitation purposes.
Training a model purely on weapon triangle weapons shows no better results than the baseline model with the optimal parameters, indicating that fewer weapon types does not improve agent performance. 

Although the model without IL implementation gives the best performance in regards of total reward, there is little guarantee that the agent imitates the player's demonstrations. Therefore, the study chooses to continue with a model combining the RL techniques from PPO and extrinsic rewards for a steady performance, together with the two IL algorithms BC and GAIL for imitation assurance. The BaselineModel+ is the final configuration, used for training the agents that mimic players in the user experiments. %The third experiment, where the models are trained based on real participant demonstrations.

% Taking the results of the first and second experiment all together, it was chosen to continue the player tests with the Mirror Model architecture computed in Experiment\ref{exp:modelconfigurations} and the hyperparameter setup found in Experiment \ref{exp:modelfinetuning}. The configuration of this model is added to the Appendix \ref{app:configurationfile}. This configuration file is used for the third experiment, where the models are trained based on real participant demonstrations.

% \section{User Tests: Mirror Mode Player Evaluation} \label{sec:playertests}

% % The model configuration found in the first two experiments discussed in Section \ref{sec:modeltesting}, is used to train the enemy agent models for the Mirror Mode. Player tests were conducted to evaluate the effect of the Mirror Mode on player experience. To evaluate this, 12 participants were found to compare their playing behavior in Standard Mode to their behavior in Mirror Mode. Participants were divided into two groups, an experimental group and a control group, each with six players.
% The final model configuration is used for agent training for Mirror Mode. The quality of the agents are evaluated through user tests. This section discusses the experimental setup for the player tests, and the collected results.

\subsection{Experiment 3: Player Studies} \label{exp:playertests}
Inspired by Pagulayan et al. \cite{pagulayan2003}, both qualitative and quantitative data were collected in a small-scale set up to evaluate the effect of the Mirror Mode algorithm on gaming experience and satisfaction. 
Only experienced strategy video game players were asked to participate. Players were asked beforehand to rate their skills and familiarity with turn-based strategy video games.

The study included 12 participants, randomly assigned to a control group ($n=6$) or an experimental group ($n=6$). The experimental group played against agents that were trained on their personal recorded demonstrations, whereas the control group played against the strategy of one player from the experimental group.
Each participant took part in two test sessions conducted on-site. In order for the AI model to learn from a participant's gameplay, the two game modes for the test group were tested on separate days.

For the first session, the participant played the Standard Mode for five full rounds. Their game state and action pairs were recorded through the player script described in \ref{subsec:agentscript}. Additional game behavior metrics were tracked in the agent script, including attack advantages and disadvantages, total movements, total attacks applied, total effective attacks, and total wins. After the playing session, a survey was administered to assess the participant's experience and satisfaction. The survey provided a satisfaction score, through 9 rating questions on a scale from 1-5, indicating a player's satisfaction of the game.

During the second session, the participant played the Mirror Mode, allowing the enemy AI to utilize the learned model from the Standard Mode. Similar playing metrics were collected, this time from the enemy agent using the Mirror Model. The session was again followed by the same survey for the satisfaction score, with additional questions that focus on comparing the Standard Mode and the Mirror Mode.

Each test was concluded with a few additional questions about the player's experience, recorded in the survey. This casual interview allowed participants to provide qualitative feedback about their overall experience across both modes. Interview questions recorded the game experience, interests, and skills of the participants. Moreover, they were asked to describe the enemy behavior and any potential differences that they noticed between the two game modes.

After the conducted user studies, a total of six distinct agent models had been trained (over a total of 1M steps): one for each experimental participant. Over all the second sessions, each model was tested twice. One time against the participant whose demonstrations were used for training the model, and one time against one control participant. This allows fair comparison of the recorded performance metrics, between the enemy agent and the participants competing against that agent. 

%part of reflection player studies results
\begin{figure*}[tbp]
    \centering

    % Row 1
    \begin{subfigure}[b]{0.48\textwidth}
        \centering
        \includegraphics[width=\textwidth]{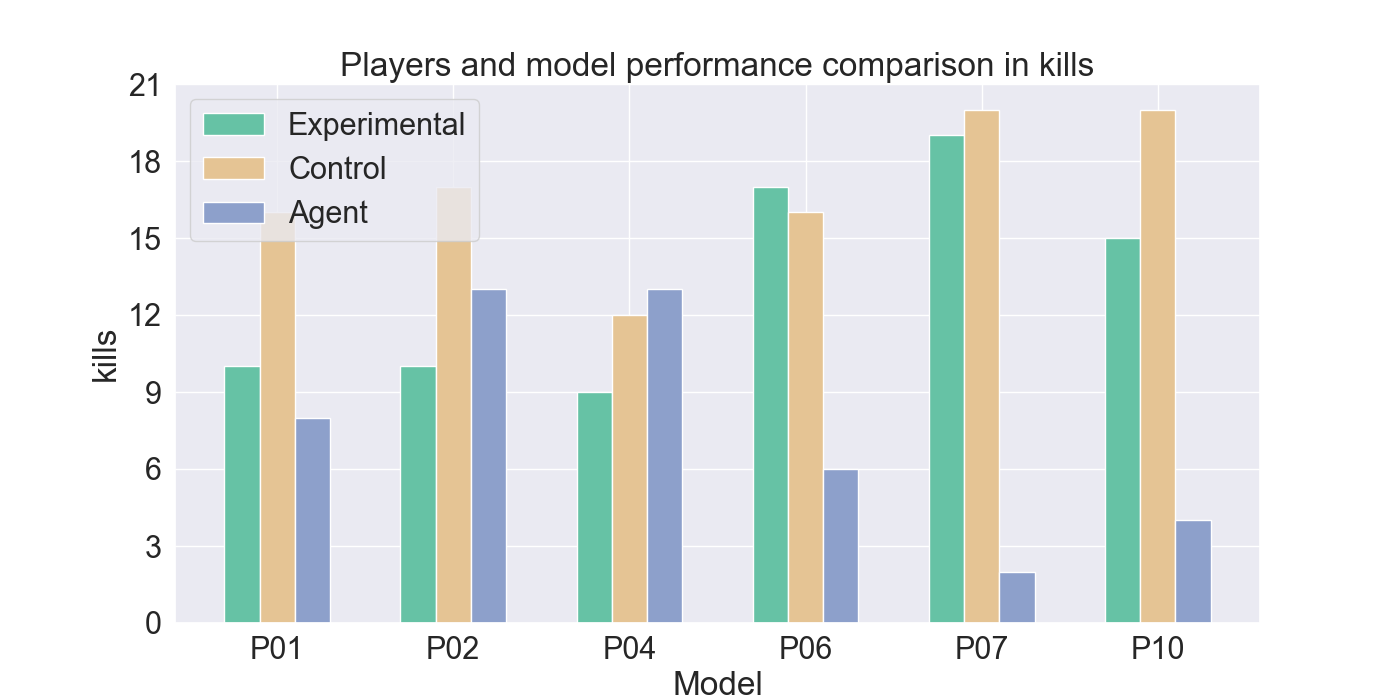}
        \caption{Total number of defeated units in the opposing team.}
        \label{fig:usertest_kills}
    \end{subfigure}
    \hfill
    \begin{subfigure}[b]{0.48\textwidth}
        \centering
        \includegraphics[width=\textwidth]{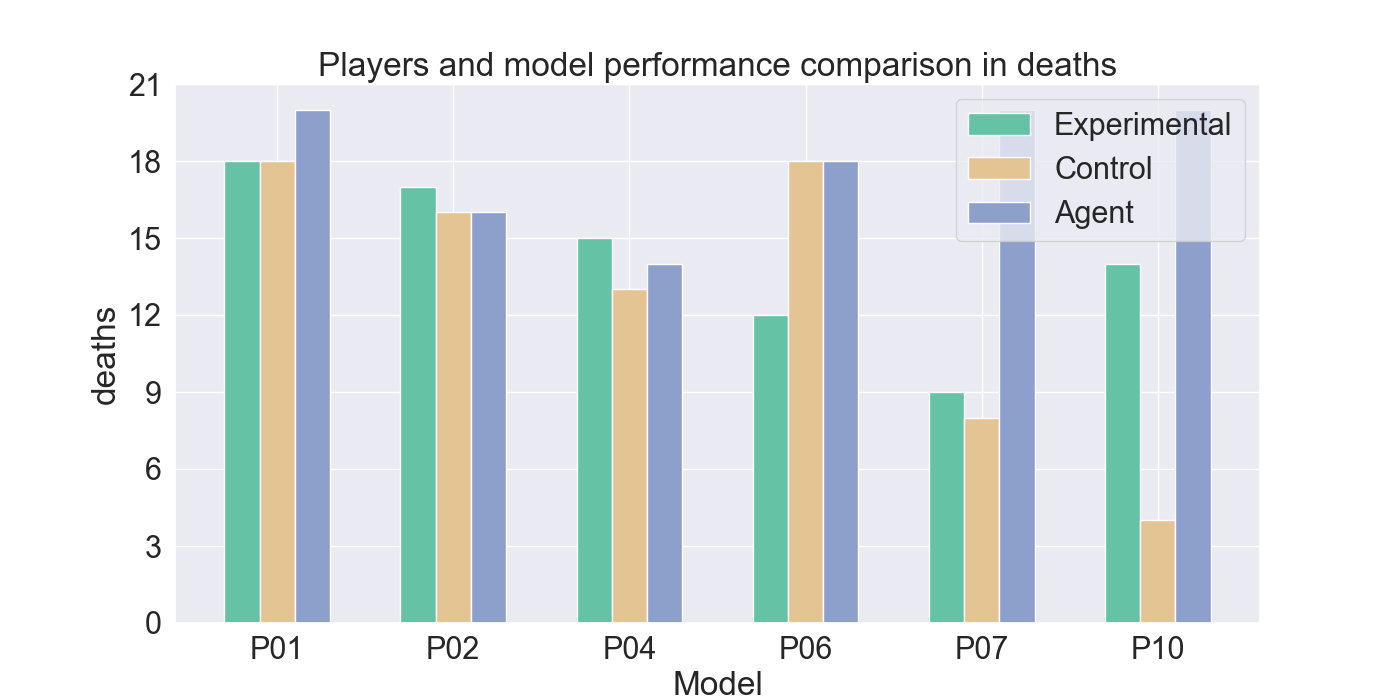}
        \caption{Total number of fallen units in the corresponding team.}
        \label{fig:usertest_deaths}
    \end{subfigure}

    \vspace{0.2em}

    % Row 2
    \begin{subfigure}[b]{0.48\textwidth}
        \centering
        \includegraphics[width=\textwidth]{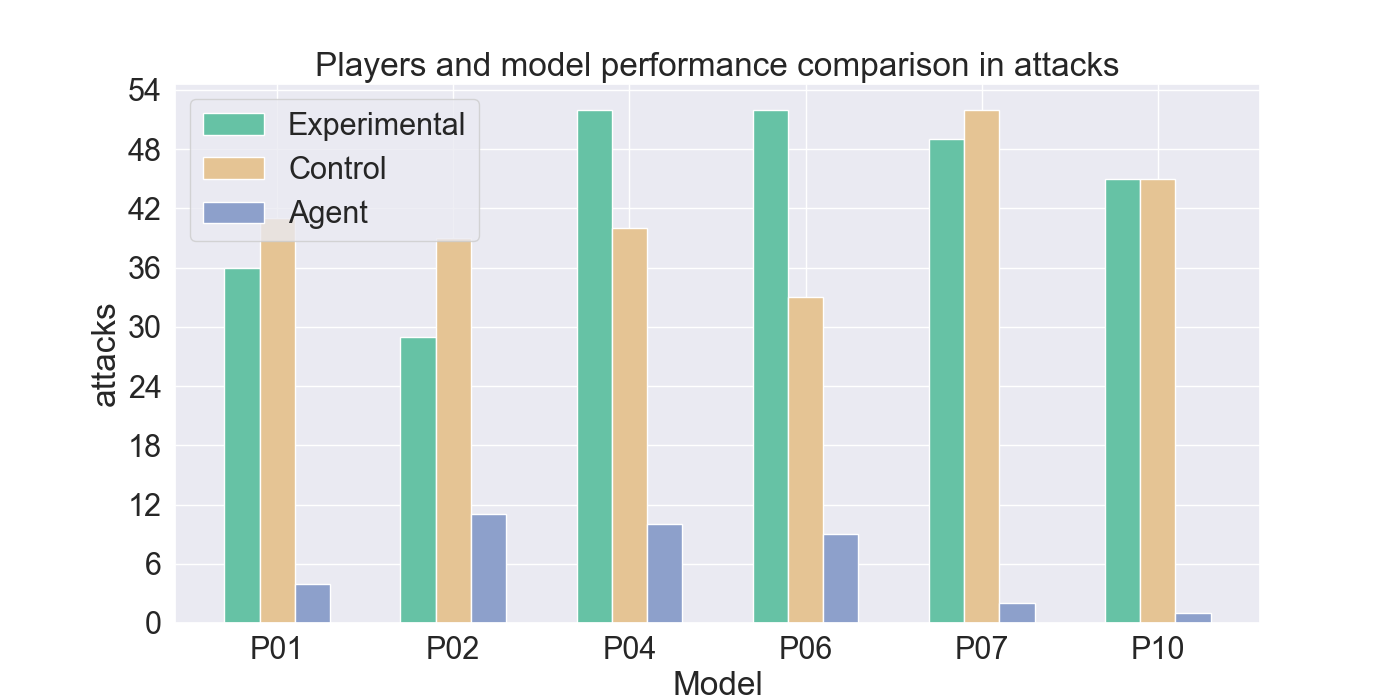}
        \caption{Total attacks performed during gameplay in a team. }
        \label{fig:usertest_attacks}
    \end{subfigure}
    \hfill
    \begin{subfigure}[b]{0.48\textwidth}
        \centering
        \includegraphics[width=\textwidth]{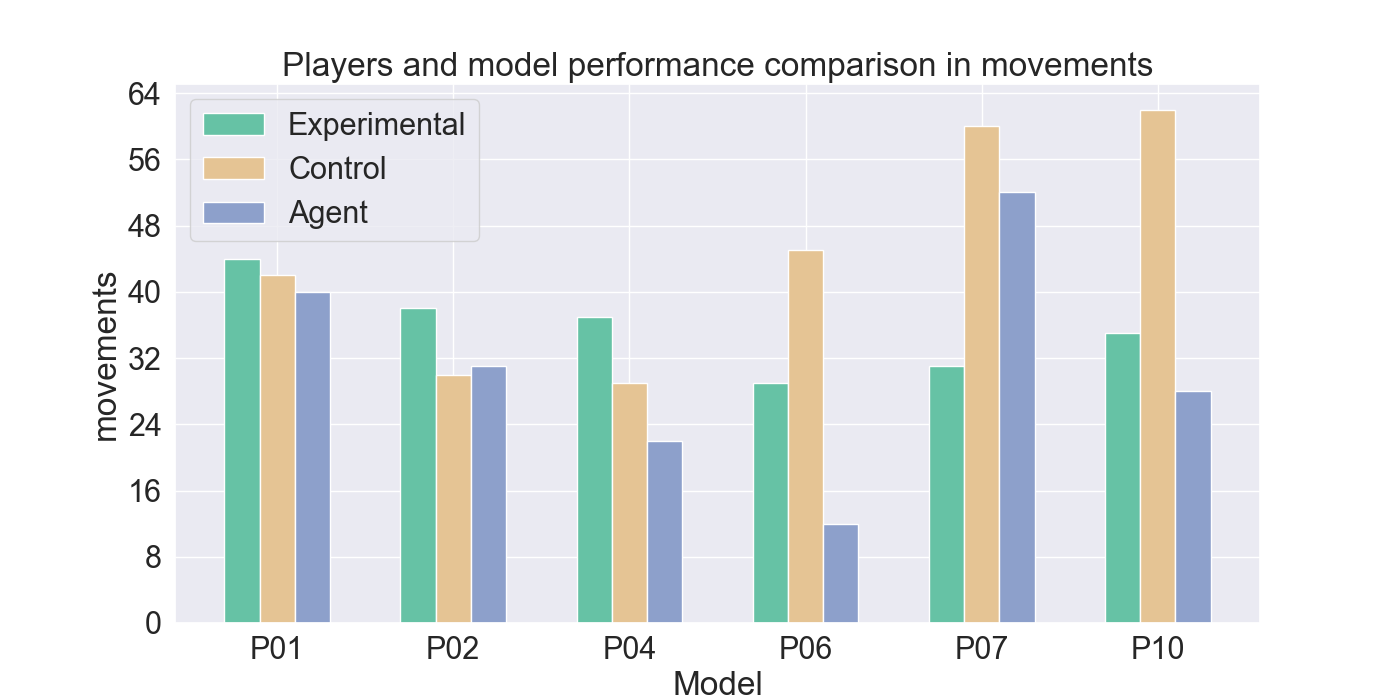}
        \caption{Total movements made by the corresponding team.}
        \label{fig:usertest_movements}
    \end{subfigure}

    \vspace{0.2em}

    % Row 3
    \begin{subfigure}[b]{0.48\textwidth}
        \centering
        \includegraphics[width=\textwidth]{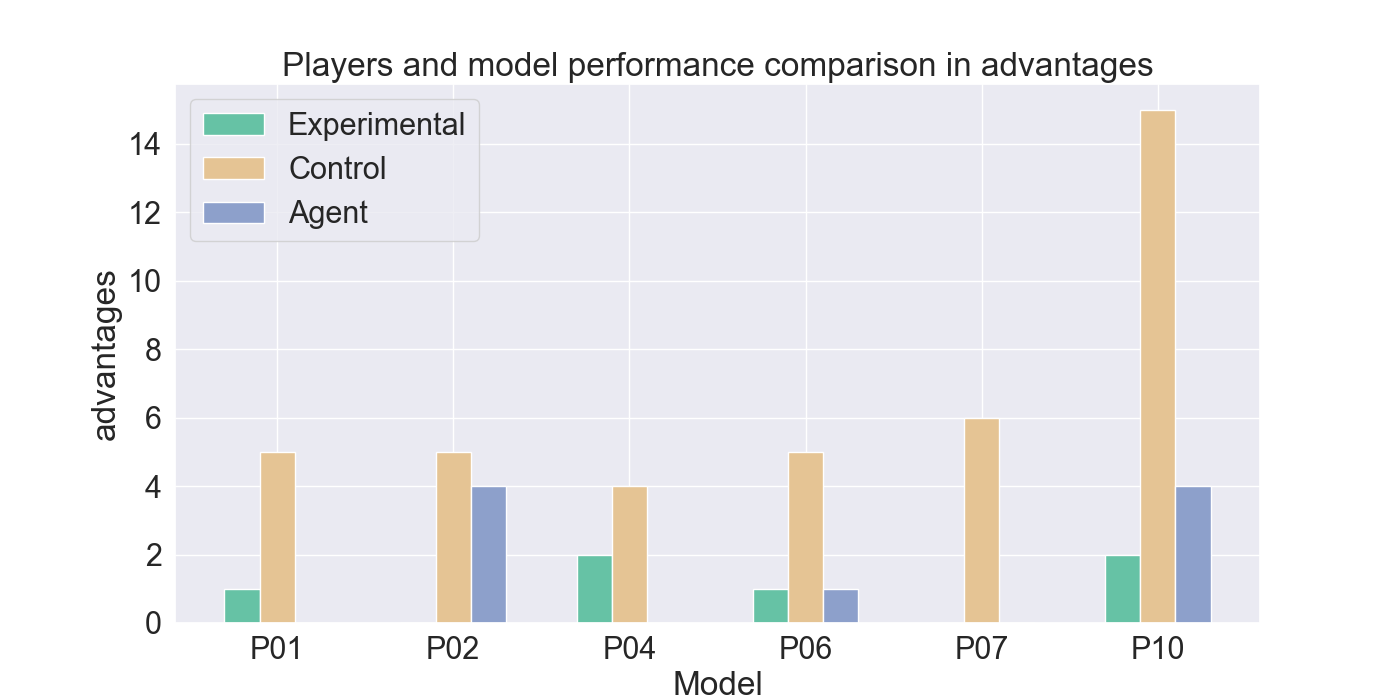}
        \caption{Total advantage attacks made by units of the measured team.}
        \label{fig:usertest_advantages}
    \end{subfigure}
    \hfill
    \begin{subfigure}[b]{0.48\textwidth}
        \centering
        \includegraphics[width=\textwidth]{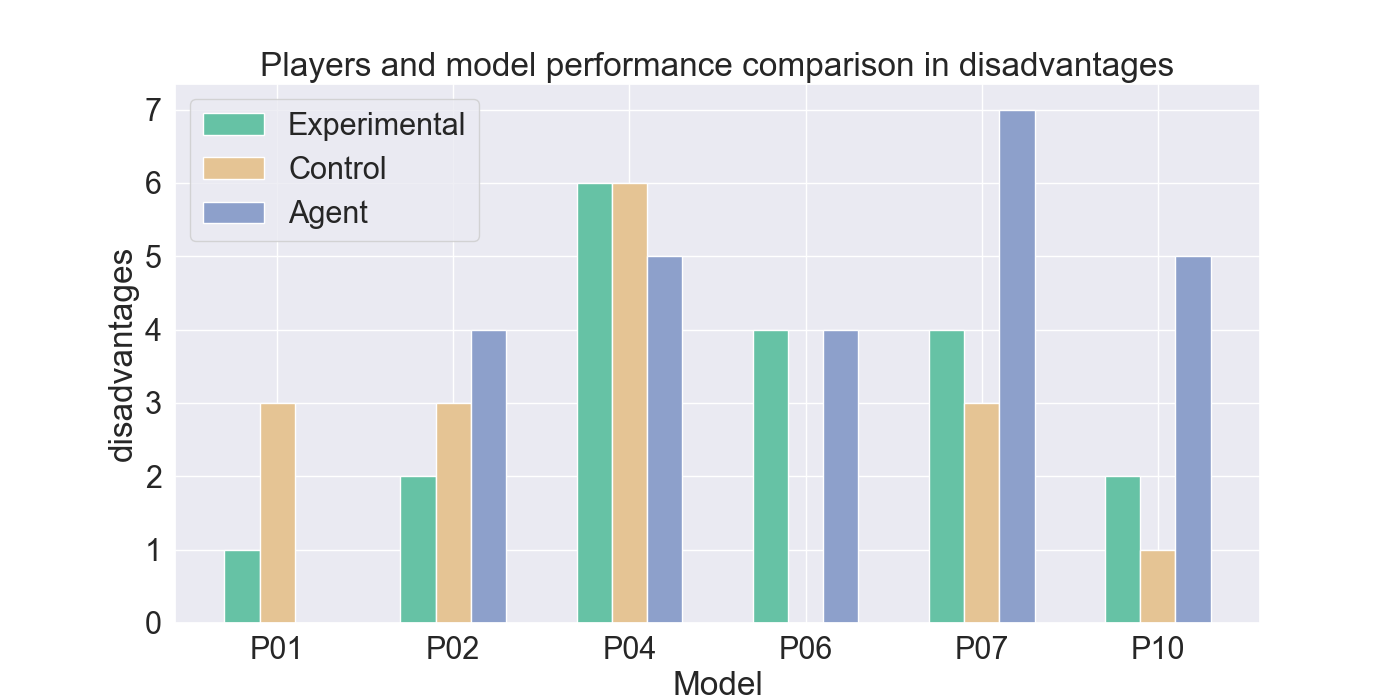}
        \caption{Reported disadvantage attacks made by units of measured team.}
        \label{fig:usertest_disadvantges}
    \end{subfigure}

    \vspace{0.2em}

    % Row 4
    \begin{subfigure}[b]{0.48\textwidth}
        \centering
        \includegraphics[width=\textwidth]{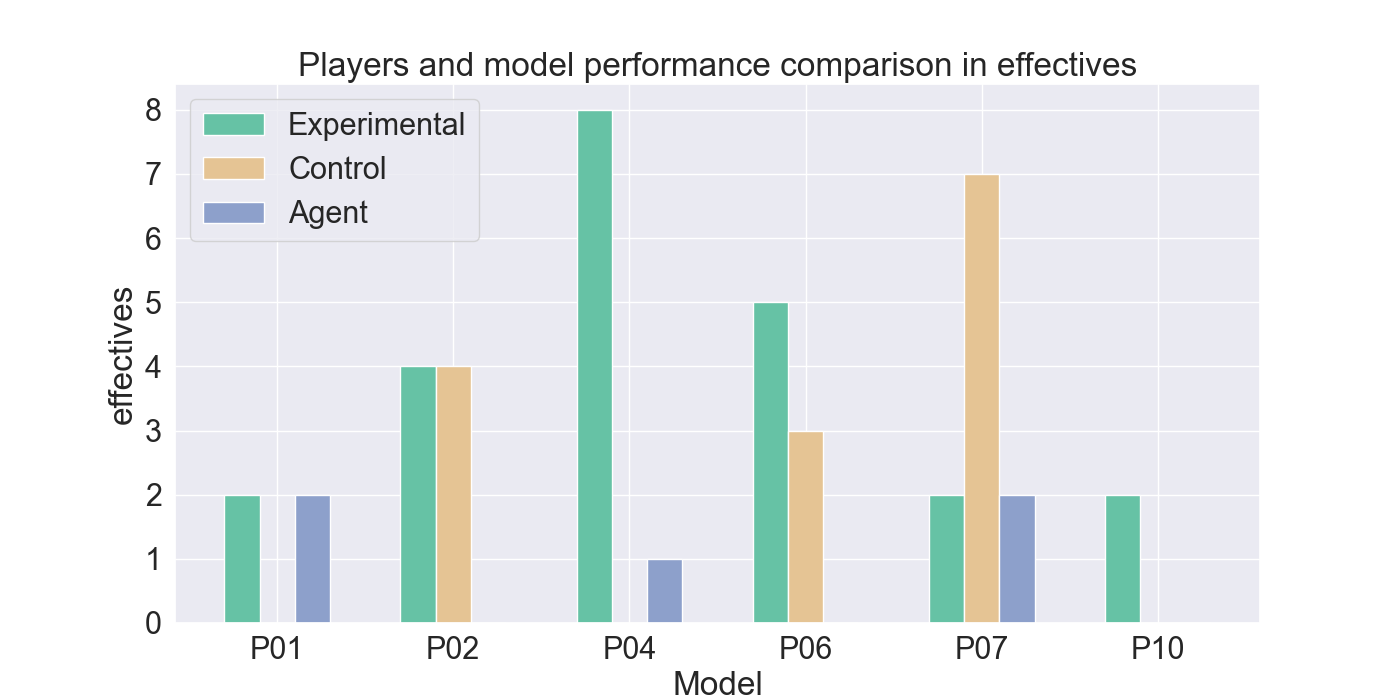}
        \caption{Number of effective attacks made by units in measured team.}
        \label{fig:usertest_effectives}
    \end{subfigure}
    \hfill
    \begin{subfigure}[b]{0.48\textwidth}
        \centering
        \includegraphics[width=\textwidth]{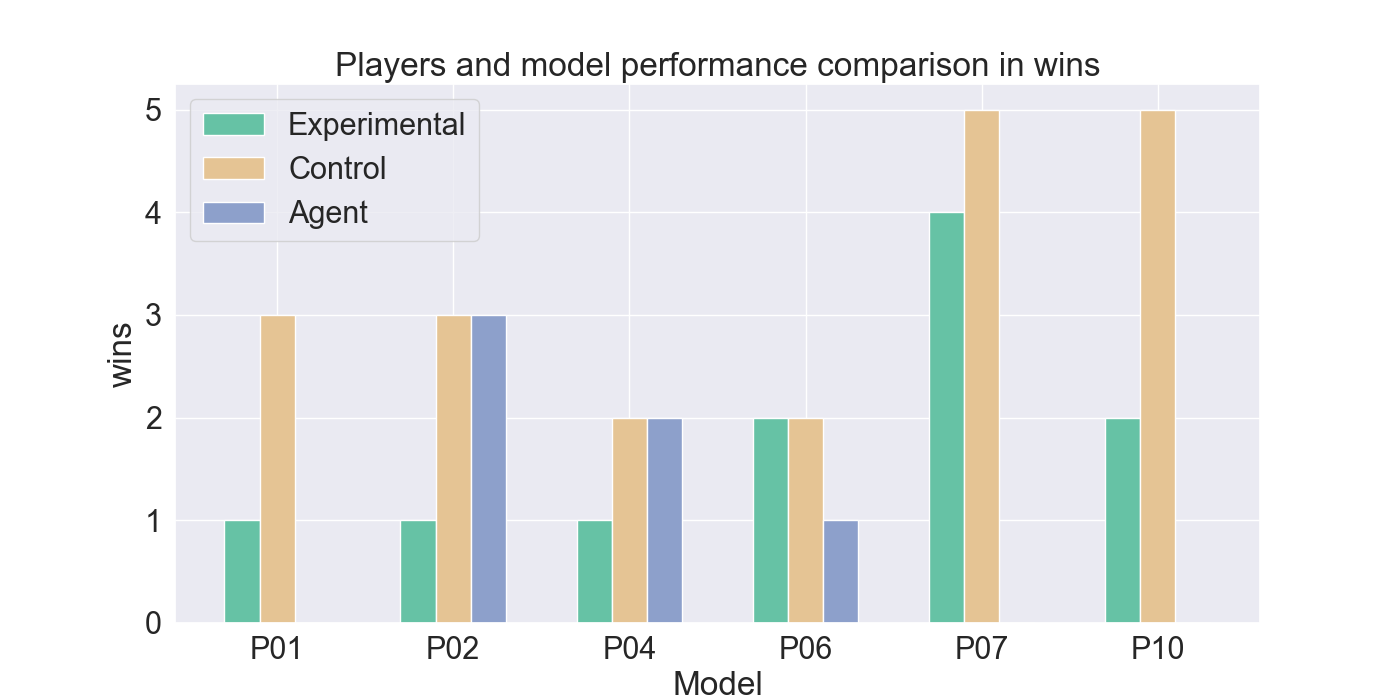}
        \caption{Number of rounds won by corresponding team.}
        \label{fig:usertest_wins}
    \end{subfigure}

    \caption{Imitation quality of the six trained agents, among measured performance metrics.}
    \label{fig:usertest_metrics}
\end{figure*}

\subsection{Reflection Player Studies Results} \label{subsec:resultsplayertests}

The results from the third experiment were collected through questionnaires, filled in by each participant. Only players familiar to strategy video games took part in this research. They were asked to rate their skills and familiarity beforehand, with the results presented in the Appendix \ref{fig:participant_skillsandexperience}. Independent t-test showed no significant differences between the experimental and control group in familiarity $t(10) = 0.00, p = 1.00$, or in skill level $t(10) = 0.42, p = 0.69$. Hence, the groups were considered fairly divided with respect to prior experience and knowledge.

\subsubsection{Player Metrics}

%In Mirror Mode, participants in the experimental group played against enemy agents that used a model trained on their own demonstrations collected during their Standard Mode gameplay. 

% The performance of the agent models is measured during Mirror Mode gameplay of each participant. The metrics of each model are taken over a total of 10 rounds. Five rounds are against its corresponding experimental participant, and five against one control participant. This way, each agent model is tested twice. The performance is based on eight in-game metrics. 
The measured performance metrics of the participants and the agent are presented in the results given in Figure \ref{fig:usertest_metrics}.
The agent performance metrics were taken in a total of 10 rounds: for both the experimental and control participant five rounds each. The total measured score was therefore divided by two, to calculate the average performance of the agent for each participant the agent competed against.  
High similarity between the agent and experimental metrics, together with low similarity to the control-group metrics, indicates strong imitation quality. Similar metric values between the control and agent bar indicate that it cannot be assured that the agent imitates the player presented by the experimental bar.

The results indicate that  Mirror Mode models struggle to imitate offensive behavior. The agents rarely perform attacks and fail to replicate effective and advantage attacks, compared to their corresponding participants. However, Figure \ref{fig:usertest_movements} shows that the agents closely resemble the movement patterns from the experimental players they were trained on. In particular, agents trained on P01, P02, and P10 demonstrate a strong alignment with their player behavior. In contrast, P04, and P06 use fewer movement actions, while P07 shows considerably more movements than its participant. 

When comparing the death rate in Figure \ref{fig:usertest_deaths} from the Mirror agents to the participants, it is noticeable that the Mirror agents show a higher death rate, than most participants. However, in most models the death rate closely matches that of the experimental participant it was trained on. Suggesting that agent's skill level is adapted to that of the corresponding player. Notably, the death rates for P06 and P07, are higher than the rates from their participants. Similarly, the kill rates for P06, P07, and P10 are substantially lower than those of their corresponding participants. 

It can be noted that the Mirror agents generally show low similarity to the control participants, supporting the idea that they are primarily imitate the strategies of the experimental group players. However, Figures \ref{fig:usertest_effectives} and \ref{fig:usertest_wins} indicate lower similarity with the experimental participants for these metrics, highlighting less accurate imitation in offensive tactics.

These observations suggest that while Mirror agents effectively replicate movement behavior from the experimental players, they lack behind in imitating offensive strategies. Nonetheless, the alignment in death rates with the original players indicates a degree of skill-level adaption in the trained models.

\subsubsection{Questionnaire}

\begin{figure}[t]
    \centering
    \includegraphics[width=1\linewidth]{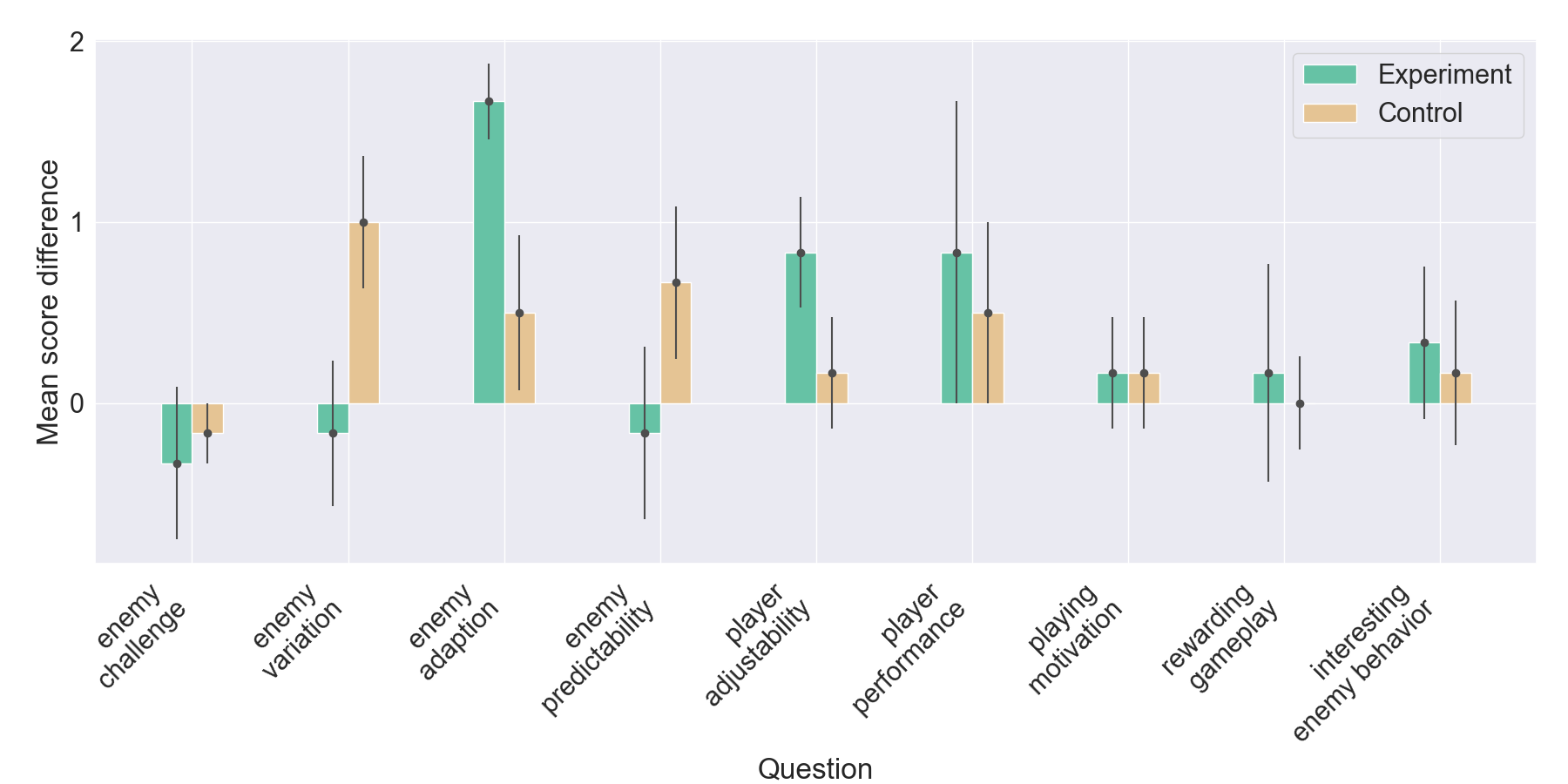}
    \caption{Mean satisfaction score difference between Mirror Mode and Standard Mode over all participants, calculated for each question. Enemy adaptation and player adjustability stand out.} %test significance van de vragen
    \label{fig:satisfaction_scores}
\end{figure}

For each of the nine rating questions, the difference in score was calculated across all participants, separately for the experimental and control groups. The mean difference in score across all participants in both groups is presented in Figure \ref{fig:satisfaction_scores}. Error bars represent the standard error of the mean (SEM), providing an indication of the precision of the estimated group means.
The y-axis lists question themes, the x-axis shows the difference in mean score, with differences calculated as $Mirror \space Mode - Standard \space Mode$. Higher values indicate greater satisfaction in Mirror Mode, while smaller SEM reflects greater confidence. Results suggest increased satisfaction in enemy adaptability and player adjustability, but a decline in perceived challenge.

Interestingly, the control group found the enemy tactics less predictable and more varied compared to experimental participants. A possible explanation for this could be that experimental participants were fighting against their own tactics, making it easier for them to predict the choices made by the enemy.
Only minor differences were observed in player's motivation to continue playing, their sense of rewarding gameplay, and their perception of enemy's behavior being interesting.
Lastly, it can be noted that the enemy challenge went down, emphasizing the observation in higher challenge of Standard Mode.

Some questions show relatively large SEM values, indicating uncertainty in the mean score differences. In particular, player performance and rewarding gameplay show a high SEM value, suggesting these scores may not accurately represent the underlying sample group. The larger error bars may partly reflect the small number of participants in each group.

\begin{figure}[t!]
    \centering
    \includegraphics[width=1\linewidth]{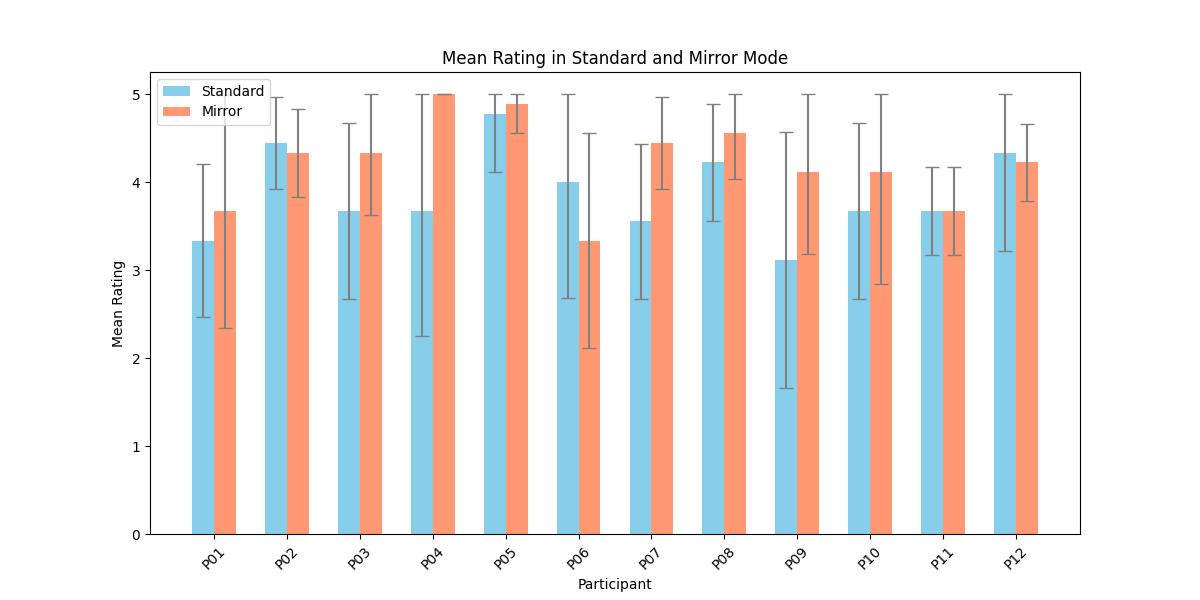}
    \caption{Mean satisfaction scores per participant; the figure illustrates that Mirror Mode scores generally higher than Standard Mode.}
    \label{fig:participant_satisfaction}
\end{figure}

\begin{table*}[tbp]
\caption{Replies to ``What differences did you notice in the enemy behavior from the Second Mode compared to the First Mode?".}
\label{tab:survey_enemydifferences}

\begin{tabular}{llp{0.8\textwidth}}
\toprule
ID  & Group & Response       \\ 
\midrule
% \rowcolor[HTML]{C0C0C0} 
P05 & B     & I noticed that the enemy behaviour changed with respect to my performance, and it was harder   than before.                                                                          \\ \hline
% \rowcolor[HTML]{C0C0C0} 
P03 & B     & I feel like the enemy backed off more often in the 2nd mode. And also that they attacked one of my people that I brought forward more often, so I could adjust my strategy better. \\ \hline
P02 & A     & The enemy behavior was more unpredictable which kept it very interesting                                                                                                             \\ \hline
% \rowcolor[HTML]{C0C0C0} 
P08 & B     & It was smarter, less greedy and taking steps more carefully. It would sometimes retreat or do surprise attacks which was unexpected.                                               \\ \hline
% \rowcolor[HTML]{C0C0C0} 
P09 & B     & Enemies would walk further away from you  making you chase. Enemies did not always go on the offensive like it seemed   in the first round.                                            \\ \hline
% \rowcolor[HTML]{C0C0C0} 
P11 & B     & Honestly i thought they were way more defensive rather than just attacking every time they could:)                                                                                     \\ \hline
P10 & A     & The enemy was staying back more, as I did.                                                                                                                                             \\ \hline
P06 & A     & Second mode used the same units as me.  Also, the magic unit was not protected as well which is a habit I have.                                                                        \\ \hline
P01 & A     & It mirrored my strategy. So it kept escaping rather than attacking.                                                                                                                    \\ \hline
P04 & A     & Played more safe and felt really dynamic   given the game state                                                                                                                        \\ \hline
P07 & A     & The biggest difference was that the enemy   decided to reposition a lot more and play more defensive, which made it a lot   more challenging to attack effectively.                    \\ \hline
% \rowcolor[HTML]{C0C0C0} 
P12 & B     & They run away more. Used the strategy I  used in the first 5 rounds but less smart executed                                                                                             \\ \hline
\bottomrule
\end{tabular}

\end{table*}

The mean of the total satisfaction scores are calculated for each participant in both Standard Mode and Mirror Mode, presented in Figure \ref{fig:participant_satisfaction}. Despite P06 and P12 giving a higher satisfaction score for Standard Mode, most participants were more satisfied in Mirror Mode compared to Standard Mode. This suggests a positive influence on their game experience by Mirror Mode. However, both modes score generally high, indicating only little effect of Mirror Mode on overall game satisfaction. 
The standard deviations are generally similar across the two modes, indicating comparable variability in responses overall and similar range of satisfaction. However, variability differs notably between participants. Participant P04 is particularly noteworthy, showing a very high standard deviation in Standard Mode but zero deviation in Mirror Mode, highlighting the increase in satisfaction for Mirror Mode.

Furthermore, participants' description of the enemy behavior differences of the two modes particularly also indicate the perception of the difference in offensive and defensive play style of the enemies, as shown in Table \ref{tab:survey_enemydifferences}. Participants in the experimental group noticed the mirroring behavior from the enemies in Mirror Mode, imitating their defensive behavior from the standard game, such as P10 who is staying back, P06 who is less protective towards the magic unit, and P01 who is more likely to retreat. Only a few in the control group recognized their own strategies in the enemy tactics in Mirror Mode, such as P05 and P12, than in the control group. 
Overall, participants reported a change in enemy strategy, moving from offensive behavior to defensive when shifting from Standard Mode to Mirror Mode.

Total satisfaction scores for each mode and participant are provided in Table \ref{tab:participant_satisfaction} in the Appendix. Comparisons of overall satisfaction between the two modes show no significant differences. A two-tailed paired t-test with $\alpha = 0.05$ resulted in $p = 0.2566$ for the experimental group and $p = 0.1144$ for the control group. 

Another significant test examined the relation between playing against one's own tactics versus another player's tactics. An independent t-test resulted in a p-value of $p = 0.9153$, indicating no significant difference between the two conditions. This surprisingly large number, suggests that what specific human demonstrations are used has minimal influence on the game experience, as long as the demonstrations likely encourage more human-like behavior.
% A summary of the computed statistical test results are presented in Table \ref{tab:satisfaction_pvalues}.
% --> iets positive van maken. 
% surprisingly it doesnt need to be trained on your own behavior, but on generic human behavior is already enough.
% --> what zou gebeuren als je zou trainen op alle data 
% \begin{table}[tbp]
% \caption{Paired samples t-test with $\alpha = 0.05$, of measured satisfaction scores between modes in the first two rows. The third row presents an independent t-test with $\alpha = 0.05$ across the two groups, comparing the difference in satisfaction score between the two modes.}
% \centering
% \begin{tabular}{lcc}
% \hline
% \textbf{Comparison} & \textbf{p-value} & \textbf{Significance} \\
% \hline
% Experimental group: \\ Standard vs Mirror Mode & 0.2566 & Not significant \\
% \hline
% Control group: \\ Standard vs Mirror Mode & 0.1144 & Not significant \\
% \hline
% Experimental vs \\ Control groups & 0.9153 & Not significant \\
% \hline
% \end{tabular}

% \label{tab:satisfaction_pvalues}
% \end{table}

\section{Discussion \& Future Work}\label{sec:discussion}
Statistics have shown a reduction in the popularity of strategy video games \cite{quantic}, potentially being caused by the repetitive and predictable behavior from NPC enemies \cite{chanel2008, lemaitre2015}. Prior studies have found advanced techniques to incorporate AI with video games, making it more enjoyable for players \cite{akram2024}. Building on this, this study introduced a new game mode in strategy video games called Mirror Mode, where the enemy agents are trained on the playing behavior of players so the agents mimic their strategy.

Even though, the experimental results demonstrate promising indications of defensive imitation in Mirror Mode, further improvements are necessary for the agents to mimic a player's overall strategy. Offensive tactics proved more challenging to replicate, with agents frequently holding back to initiate attacks or failing to exploit advantageous or effective actions. There are several reasons likely causing this behavior:
\begin{itemize}
    \item Learning to attack properly requires the agent to learn to recognize when an opponent can be attacked, and from which tile they can attack that chosen opponent. This gives two additional steps compared to the singular act to move to a different tile. 
    \item The standard enemy AI had a high challenge, causing participants to struggle to defeat them and forcing them to retreat. As observed, players showed a higher death rate in Standard Mode, and rated the enemy AI as more challenging. This suggests a potential domino effect, where players who performed suboptimally in demonstrations, inadvertently transferred those struggles to their agents. Consequently, the agents inherited and reproduced similar results during training and their final model.
    \item The models were trained for 1M steps. This may have not given enough time for the agent to learn proper attacking behavior.
\end{itemize}

Future work should focus on improving the game to decrease the challenge from the Standard Mode by incorporating more elements from the original games, such as abilities. Ideally, a testing environment within the real mobile game of Fire Emblem Heroes could develop more insights, reaching a larger audience, and enable real-time gameplay over an extended period of gameplays.
This would allow the model to continuously integrate new demonstrations and adjusts to new player strategies that involve counter strategies developed against their own behavior.
We also recommend a larger user study in order to achieve statistically significant results more easily.

Moreover, in-game evaluation can be used to assess imitation quality of different configured models, for real game performance and imitation insights, to find the most optimal model configuration.
Refinements to the reward system are also necessary. In particular, providing explicit rewards for replication actions from recorded demonstrations. This could further improve imitation performance.

% Ultimately, better performing agents, together with a larger user group in future work could increase the satisfaction of participants significantly. As due to an insufficient sample size and implementation, the satisfaction between the two modes was insignificantly different.    

\section{Conclusion}\label{sec:conclusion}

This research introduces a new gameplay mode called Mirror Mode, for turn-based strategy games. It is found that RL and IL techniques from the Unity ML-Agent package provided by Juliani et al. \cite{juliani2020} show big potentials for teaching agents to copy the strategy of a player, in particular defensive tactics. 

The first half of the study focused on providing a model for Mirror Mode. Leading to an answer to the question: ``\textit{To what extent can RL and IL be applied to teach NPCs a player's strategy in a turn-based strategy game?}'' 
It can be concluded that IL has a strong potential in teaching enemy agents a player's strategy, with the proper configurations to maintain the imitation-exploration trade-off. RL is more adaptable in finding its own strategy, and less suitable for copying player's strategies. Therefore, IL is more preferable for the specific purpose of teaching NPCs a player's strategy, with PPO and extrinsic rewards ensuring a good agent performance.

The second half of the study provided insights on the imitation quality in Mirror Mode, and the experience when playing this mode. Game metrics were taken from the participant's gameplay, indicating a good imitation in defensive behavior rather than offensive tactics. Participants recognized their own retreating tactics in the Mirror Mode enemies. 
Surveys were taken to rate the participant's game satisfaction for both modes, resulting in an overall higher satisfaction for Mirror Mode, though not significant yet.

In addressing the general question
``\textit{How will a player's game experience be influenced when NPCs imitate their strategy in a turn-based strategy game?}'', this study therefore shows that the overall satisfaction of the game is moderately increased. Participants enjoyed Mirror Mode better due to the less predictable enemy behavior and their recognized defensive strategies, but making them easier to defeat. 

In conclusion, Mirror Mode can increase the satisfaction of players in strategy games. Whether this can increase the popularity of  strategy games remains to be discussed, but this study takes a first step. 

\bibliographystyle{ACM-Reference-Format}
\bibliography{sources}

@misc{myproject,
  author = {Anonymous},
  title = {Mirror Mode Research},
  year = {2025},
  publisher = {GitHub},
  journal = {GitHub repository},
    url = {https://github.com/AnonymousResearcher22/MirrorModeResearch},
}

@misc{ML-Agent,
    author  = {{Unity Technologies}},
    title   = {{ML-Agents Overview}},
    year    = {2024},
    journal = {Unity},
    url     = {https://docs.unity3d.com/Packages/com.unity.ml-agents@3.0/manual/index.html}
}

@article{unity-ml-agents-parameters,
  title={{Unity ML-Agents Toolkit}},
  author={{Unity Technologies}},
  journal={The Unity Machine Learning Agents Toolkit (ML-Agents)},
  year={2022},
  publisher={Github}, 
  url={https://unity-technologies.github.io/ml-agents/Training-Configuration-File/#gail-intrinsic-reward}
}

@article{ML-Agent-Installation,
    author  ={Miguel Alonso Jr.},
    title   = {{ML-Agents Installation}},
    year    = {2024},
    journal = {The Unity Machine Learning Agents Toolkit (ML-Agents)},
publisher = {Github},
}

@article{ML-Agent-Reward,
    author  = {{Unity Technologies}},
    title   = "Reward Signals",
    year    = "2020",
    journal = "The Unity Machine Learning Agents Toolkit (ML-Agents)",
    url     = "https://github.com/Unity-Technologies/ml-agents/blob/0.15.0/docs/Reward-Signals.md"
}

@article{juliani2020,
  title={Unity: A general platform for intelligent agents},
  author={Juliani, Arthur and Berges, Vincent-Pierre and Teng, Ervin and Cohen, Andrew and Harper, Jonathan and Elion, Chris and Goy, Chris and Gao, Yuan and Henry, Hunter and Mattar, Marwan and Lange, Danny},
  journal={arXiv preprint arXiv:1809.02627},
  doi={https://doi.org/10.48550/arXiv.1809.02627},
  year={2020}
}

@article{enemyAI,
      title={{How Does Enemy AI Work?}}, 
      author={{Fire Emblem Heroes (FEH) Walkthrough Team}},
      year={2021},
      journal={Game8},
      url={https://game8.co/games/fire-emblem-heroes/archives/324532}, 
}

@article{FEUIsprites,
    author  = {{Fire Emblem Heroes Wiki}},
    title   = "Game Assets Collection",
    journal = {Fandom Games Community},
    year    = "2025",
    url     = "https://feheroes.fandom.com/wiki/Game_assets_collection#UI_Sprite_sheets"
}

@article{FEherolist,
    author  = {Fire Emblem Heroes Wiki},
    title   = "List of Heroes",
    journal = "Fandom Games Community",
    year    = "2025",
    url     = "https://feheroes.fandom.com/wiki/List_of_Heroes"
}

@article{NintendoFEH,
    author = "Intelligent Systems",
    title = "Fire Emblem Heroes",
    journal = "Nintendo",
    year = "2017"
}

@article{ross2010,
  title = 	 {{Efficient Reductions for Imitation Learning}},
  author = 	 {Ross, Stephane and Bagnell, Drew},
  journal = 	 {{Proceedings of the Thirteenth International Conference on Artificial Intelligence and Statistics  (AISTATS)}},
  pages = 	 {661--668},
  year = 	 {2010},
  editor = 	 {Teh, Yee Whye and Titterington, Mike},
  volume = 	 {9},
  series = 	 {Proceedings of Machine Learning Research},
  address = 	 {Chia Laguna Resort, Sardinia, Italy},
  month = 	 {13--15 May},
  publisher =    {PMLR},
  url = 	 {https://proceedings.mlr.press/v9/ross10a.html},
}

@misc{amatoandshani,
    author = {Amato, Christopher and Shani, Guy},
    year = {2010},
    month = {01},
    pages = {75-82},
    title = {{High-level reinforcement learning in strategy games}},
    volume = {1},
    journal = {{Proceedings of the International Joint Conference on Autonomous Agents and Multiagent Systems, AAMAS}},
    doi = {10.1145/1838206.1838217}
}

@misc{baker2020,
      title={{Emergent Tool Use From Multi-Agent Autocurricula}}, 
      author={Bowen Baker and Ingmar Kanitscheider and Todor Markov and Yi Wu and Glenn Powell and Bob McGrew and Igor Mordatch},
      year={2020},
      eprint={1909.07528},
      archivePrefix={arXiv},
      primaryClass={cs.LG},
      doi = {https://doi.org/10.48550/arXiv.1909.07528}
    
}

@article{openai2019dota,
  title={{Dota 2 with Large Scale Deep Reinforcement Learning}},
  author={OpenAI and Christopher Berner and Greg Brockman and Brooke Chan and Vicki Cheung and Przemysław Dębiak and Christy Dennison and David Farhi and Quirin Fischer and Shariq Hashme and Chris Hesse and Rafal Józefowicz and Scott Gray and Catherine Olsson and Jakub Pachocki and Michael Petrov and Henrique Pondé de Oliveira Pinto and Jonathan Raiman and Tim Salimans and Jeremy Schlatter and Jonas Schneider and Szymon Sidor and Ilya Sutskever and Jie Tang and Filip Wolski and Susan Zhang},
  year={2019},
  eprint={1912.06680},
  archivePrefix={arXiv},
  doi={
https://doi.org/10.48550/arXiv.1912.06680}
}

@article{gharbi2024,
    author = {Gharbi, Hafsa and Fennan, Abdelhadi and Lotfi, Elaachak},
    year = {2024},
    month = {08},
    pages = {5735},
    title = "Replicating video game players’ behavior through {Deep Reinforcement Learning} algorithms",
    volume = {102},
    journal = {Journal of Theoretical and Applied Information Technology}
}

@misc{HOandERMONGAIL,
author = {Ho, Jonathan and Ermon, Stefano},
year = {2016},
month = {06},
pages = {},
title = {{Generative Adversarial Imitation Learning}},
doi = {10.48550/arXiv.1606.03476}
}

@misc{sanchezruiz2008,
author = {Sánchez-Ruiz, Antonio and Stephen, Ruiz and Héctor, Lee-Urban and Noz-Avila, M and Díaz Agudo, Belen},
year = {2008},
month = {07},
pages = {},
title = {{Game AI for a Turn-based Strategy Game with Plan Adaptation and Ontology-based retrieval}}
}

@misc{Fan2023,
  title={The Application of Reinforcement Learning in Video Games},
  author={Xuyouyang Fan},
  year={2023},
  booktitle={Proceedings of the 2023 International Conference on Image, Algorithms and Artificial Intelligence (ICIAAI 2023)},
  pages={202-211},
  issn={2352-538X},
  isbn={978-94-6463-300-9},
  url={https://doi.org/10.2991/978-94-6463-300-9_21},
  doi={10.2991/978-94-6463-300-9_21},
  publisher={Atlantis Press}
}

@misc{zare2023,
      title={{A Survey of Imitation Learning: Algorithms, Recent Developments, and Challenges}}, 
      author={Maryam Zare and Parham M. Kebria and Abbas Khosravi and Saeid Nahavandi},
      year={2023},
      eprint={2309.02473},
      archivePrefix={arXiv},
      primaryClass={cs.LG},
      doi={
https://doi.org/10.48550/arXiv.2309.02473}
}

@misc{lemaitre2015,
    author = {Lemaitre, Juliette and Lourdeaux, Domitile and Chopinaud, Caroline},
    year = {2015},
    month = {01},
    pages = {},
    title = {{Towards a Resource-based Model of Strategy to Help Designing Opponent AI in RTS Games}},
    volume = {1},
    doi = {10.5220/0005254402100215}
}

@article{tian2024,
author = {Tian, Xinhe},
year = {2024},
month = {03},
pages = {161-170},
title = {AI applications in video games and future expectations},
volume = {54},
journal = {Applied and Computational Engineering},
doi = {10.54254/2755-2721/54/20241484}
}

@misc{chang2020,
    author = {Chang, SC. and Chiu, YP. and Hwang, JC.},
    year = {2020},
    month = {03},
    pages = {369–382},
    title = {{Determining Satisfaction from Gameplay by Discussing Flow States Related to Relaxation and Excitement}},
    journal = {{The Computer Games Journal}},
    volume = {9},
    doi = {https://doi.org/10.1007/s40869-020-00113-5}
}

@article{huang2023,
author = {Huang, Richard},
year = {2023},
month = {09},
pages = {43-48},
title = {{The Impact of Flow State and Immersion in Video Games}},
volume = {5},
journal = {Communications in Humanities Research},
doi = {10.54254/2753-7064/5/20230028}
}

@inproceedings{chanel2008,
author = {Chanel, Guillaume and Rebetez, Cyril and B\'{e}trancourt, Mireille and Pun, Thierry},
title = {Boredom, engagement and anxiety as indicators for adaptation to difficulty in games},
year = {2008},
isbn = {9781605581972},
publisher = {Association for Computing Machinery},
address = {New York, NY, USA},
url = {https://doi.org/10.1145/1457199.1457203},
doi = {10.1145/1457199.1457203},
booktitle = {Proceedings of the 12th International Conference on Entertainment and Media in the Ubiquitous Era},
pages = {13–17},
numpages = {5},
location = {Tampere, Finland},
series = {MindTrek '08}
}

@misc{quantic,
    author = {Nick Yee},
    year = {2024},
    month = {05},
    title = {Gamers Have Become Less Interested in Strategic Thinking and Planning},
    journal = {Quantic Foundry},
    url = {https://quanticfoundry.com/2024/05/21/strategy-decline/}
}

@book{csíkszentmihályi,
    author = {Mihály Csikszentmihalyi},
    year = {1975},
    title = {Beyond Boredom and Anxiety},
    publisher = {Jossey-Bass Publishers}
}

@article{akram2024,
    author = {Akram, Arslan and Tehseen, Rabia and Saqib, Shazia and Nazir, Faria and Awan, Maham and Jr, Ijist},
    year = {2024},
    month = {12},
    pages = {2004-2023},
    title = {{Advanced AI Mechanics in Unity 3D for Immersive Gameplay. A Study on Finite State Machines \& Artificial Intelligence}},
    volume = {6},
    journal ={ {International Journal of Innovations in Science and Technology}}
}

@article{pagulayan2003,
author = {Pagulayan, Randy and Keeker, Kevin and Wixon, Dennis and Romero, Ramon and Fuller, Thomas},
year = {2003},
month = {01},
pages = {883-906},
title = {{User-Centered Design in Games}},
isbn = {0-8058-3838-4},
journal ={ {Human-Computer Interact. Handb}},
doi = {10.1201/b11963-39}
}

@article{lee2012,
   author = "Lee, Michael Sangyeob and Heeter, Carrie",
   title = "What do you mean by believable characters?: The effect of character rating and hostility on the perception of character believability", 
   journal= "Journal of Gaming \&amp; Virtual Worlds",
   year = "2012",
   volume = "4",
   number = "1",
   pages = "81-97",
   doi = "https://doi.org/10.1386/jgvw.4.1.81_1",
   url = "https://intellectdiscover.com/content/journals/10.1386/jgvw.4.1.81_1",
   publisher = "Intellect",
   issn = "1757-1928",
  
  }

\appendix

\section{Participant Experience and Skills Information}
\begin{figure}[h]
    \centering
    \includegraphics[width=0.9\linewidth]{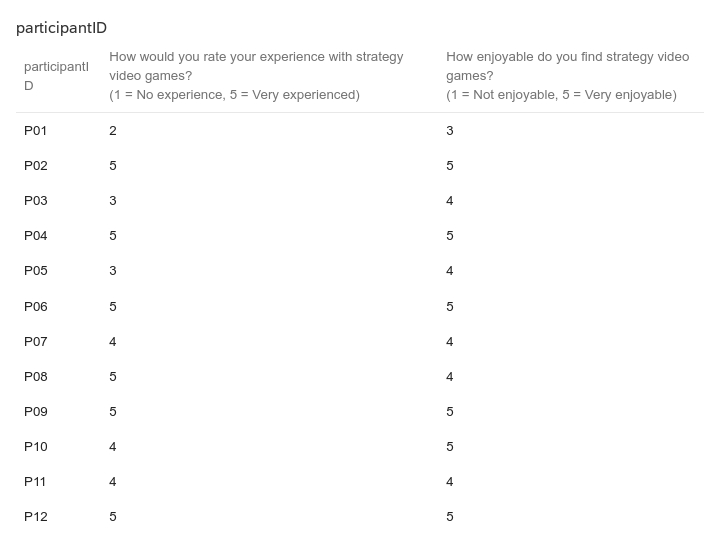}
    \caption{Experience and skill level of each participant, rated by the participants themselves.}
    \label{fig:participant_skillsandexperience}
\end{figure}

%\newpage

\section{Total Satisfaction Scores Standard Mode and Mirror Mode}\label{tab:participant_satisfaction}
\begin{table}[H]
\centering
\begin{tabular}{l|l|l|l|l}
ID  & Group & Standard Mode & Mirror Mode & Score Difference \\
P01 & A     & 30       & 33     & 3     \\
P02 & A     & 40       & 39     & -1     \\
P03 & B     & 33       & 39     & 6     \\
P04 & A     & 33       & 45     & 12     \\
P05 & B     & 43       & 44     & 1     \\
P06 & A     & 36       & 30     & -6     \\
P07 & A     & 32       & 40     & 8     \\
P08 & B     & 38       & 41     & 3     \\
P09 & B     & 28       & 37     & 9       \\
P10 & A     & 33       & 37     & 4     \\
P11 & B     & 33       & 33     & 0     \\
P12 & B     & 39       & 38     & -1   
\end{tabular}
\caption{Uitleg tabel. Mirror Mode scores generally higher than Standard Mode}%distributie van score differences
% naar github
% clean github maken

\end{table}

\end{document}